\documentclass{article} % For LaTeX2e
\usepackage{iclr2023_conference,times}

\usepackage{amsmath,amsfonts}

\def\eqref#1{equation~\ref{#1}}
\def\1{\bm{1}}

\DeclareMathAlphabet{\mathsfit}{\encodingdefault}{\sfdefault}{m}{sl}
\SetMathAlphabet{\mathsfit}{bold}{\encodingdefault}{\sfdefault}{bx}{n}

\usepackage{hyperref}
\usepackage{booktabs}       % professional-quality tables
\usepackage{amsfonts}       % blackboard math symbols
\usepackage{nicefrac}       % compact symbols for 1/2, etc.
\usepackage{microtype}      % microtypography
\usepackage{subcaption}
\usepackage{graphicx}
\usepackage{amssymb}
\usepackage{amsthm}
\usepackage{mathtools}
\usepackage{wrapfig}
\usepackage{enumitem}
\usepackage{makecell, cellspace, caption}
\usepackage{soul}

\newtheorem{proposition}{Proposition}

 \title{Learning Topology-Preserving Data Representations}

 \author{Ilya Trofimov$^1$, Daniil Cherniavskii$^{1,4}$, Eduard Tulchinskii$^{1,3}$, Nikita Balabin$^1$,\\ \textbf{Evgeny Burnaev}\thanks{Equal senior contribution.}$^{\,\,\,,1,4}$,
 ~\textbf{Serguei Barannikov}$^{*, 1,2}$\\ $^1$Skolkovo Institute of Science and Technology;\\\textsuperscript{2}CNRS, Université Paris Cité; \textsuperscript{3}Huawei Noah’s Ark lab; 
\\ $^4$Artificial Intelligence Research Institute (AIRI)}

\iclrfinalcopy % Uncomment for camera-ready version, but NOT for submission.
\begin{document}

\maketitle

\begin{abstract}
We propose a method for learning topology-preserving data representations (dimensionality reduction). The method aims to provide topological similarity between the data manifold and its latent representation via enforcing the similarity in topological features (clusters,  loops, 2D voids, etc.) and their localization. 
The core of the method is the minimization of the Representation Topology Divergence (RTD) between original high-dimensional data and low-dimensional representation in latent space. RTD minimization provides closeness in topological features with strong theoretical guarantees. 
We develop a scheme for RTD differentiation and apply it as a loss term for the autoencoder. The proposed method ``RTD-AE'' better preserves the global structure and topology of the data manifold than state-of-the-art competitors as measured by linear correlation, triplet distance ranking accuracy, and Wasserstein distance between persistence barcodes. \let\thefootnote\relax\footnotetext{Correspondence: i-tr@yandex.ru}\end{abstract}

\section{Introduction}
\label{sec:intro}

Dimensionality reduction is a useful tool for data visualization, preprocessing, and exploratory data analysis. Clearly, immersion of high-dimensional data into 2D or 3D space is impossible without distortions which vary for popular methods. Dimensionality reduction methods can be broadly classified into global and local methods. Classical global methods (PCA, MDS) tend to preserve the global structure of a manifold. However, in many practical applications, produced visualizations are non-informative since they don't capture complex non-linear structures.
Local methods (UMAP \citep{mcinnes2018umap}, PaCMAP \citep{wang2021understanding}, t-SNE \citep{van2008visualizing},
Laplacian Eigenmaps \citep{belkin2001laplacian}, ISOMAP \citep{tenenbaum2000global}) focus on preserving neighborhood data and local structure with the cost of sacrificing the global structure. The most popular methods like t-SNE and UMAP are a good choice for inferring cluster structures but often fail to describe correctly the data manifold's topology. t-SNE and UMAP have hyperparameters influencing representations neighborhood size taken into account. Different values of hyperparameters lead to significantly different visualizations and neither of them is the ``canonical'' one that correctly represents high-dimensional data.

We take a different perspective on dimensionality reduction.
We propose the approach based on \textit{Topological Data Analysis (TDA)}. Topological Data Analysis \citep{Barannikov1994,zomorodian2001computing,chazal2017introduction}
is a field devoted to the numerical description of multi-scale topological properties of data distributions by analyzing point clouds sampled from them. TDA methods naturally capture properties of data manifolds on multiple distance scales and are arguably a good trade-off between local and global approaches.

The state-of-the-art TDA approach of this kind is TopoAE \citep{moor2020topological}. However, it has several weaknesses: 1) the loss term is not continuous 2) the nullity of the loss term is only necessary but not a sufficient condition for the coincidence of topology, as measured by persistence barcodes, see more details in Appendix \ref{app:topoae_critique}.

In our paper, we suggest using the Representation Topology Divergence (RTD) \citep{barannikov2021representation} to produce topology-aware dimensionality reduction. RTD measures the topological discrepancy between two point clouds with one-to-one correspondence between clouds and enjoys nice theoretical properties (Section \ref{sec:rtd}). The major obstacle to incorporate RTD into deep learning is its differentiation. There exist approaches to the differentiation of barcodes, generic barcodes-based functions with respect to deformations of filtration \citep{carriere2021optimizing} and to TDA differentiation in special cases \citep{hofer2019connectivity, poulenard2018topological}.

In this paper, we make the following contributions:
%\begin{enumerate}[topsep=0pt,noitemsep,nolistsep, partopsep=0pt, parsep=0ex, leftmargin=*]
%\begin{enumerate}
\begin{enumerate}[topsep=0pt,noitemsep,nolistsep, partopsep=0pt, parsep=0ex, leftmargin=*]
    \item We develop an approach for RTD differentiation. Topological metrics are difficult to differentiate; the differentiability of RTD and its implementation on GPU is a valuable step forward in the TDA context which opens novel possibilities in topological optimizations;
    \item We propose a new method for topology-aware dimensionality reduction: an autoencoder enhanced with the differentiable RTD loss: ``RTD-AE''. Minimization of RTD loss between real and latent spaces forces closeness in topological features and their localization  with strong theoretical guarantees;
    \item By doing computational experiments, we show that the proposed RTD-AE outperforms state-of-the-art methods of dimensionality reduction and the vanilla autoencoder in terms of preserving the global structure and topology of a data manifold; we measure it by the linear correlation, the triplet distance ranking accuracy, Wasserstein distance between persistence barcodes, and RTD. In some cases, the proposed RTD-AE produces more faithful and visually appealing low-dimensional embeddings than state-of-the-art algorithms. We release the RTD-AE source code. \footnote{\href{https://github.com/danchern97/RTD_AE}{github.com/danchern97/RTD\_AE}}
\end{enumerate}

%
%   Teaser
%
\begin{figure}[t]
    \centering
    \begin{subfigure}{0.16\textwidth}
    \includegraphics[width=\textwidth]{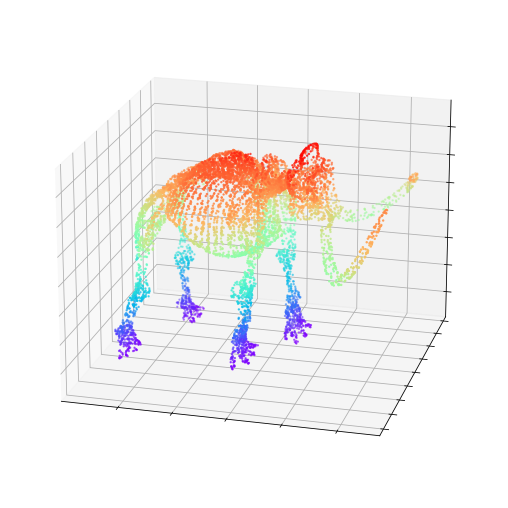}
    \caption{Orig. 3D data}
    \end{subfigure}
    \begin{subfigure}{0.16\textwidth}
    \includegraphics[width=\textwidth]{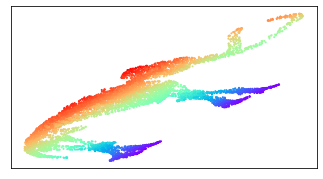}
    \caption{Autoencoder}
    \end{subfigure}
    \begin{subfigure}{0.16\textwidth}
    \includegraphics[width=\textwidth]{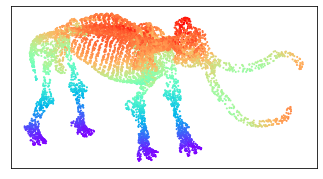}
    \caption{RTD-AE}
    \end{subfigure}
    \begin{subfigure}{0.16\textwidth}
    \includegraphics[width=\textwidth]{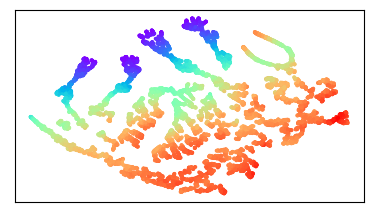}
    \caption{t-SNE}
    \end{subfigure}
    \begin{subfigure}{0.16\textwidth}
    \includegraphics[width=\textwidth]{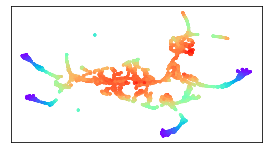}
    \caption{UMAP}
    \end{subfigure}
    \begin{subfigure}{0.16\textwidth}
    \includegraphics[width=\textwidth]{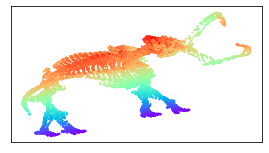}
    \caption{TopoAE}
    \end{subfigure}
    \caption{Dimensionality reduction (3D $\to$ 2D) on the ``Mammoth'' dataset. The proposed RTD-AE method better captures both global and local structure.}
    \label{fig:mammoth}
\end{figure}

\section{Related work}

Various dimensionality reduction methods have been proposed to obtain 2D/3D visualization of high-dimensional data \citep{tenenbaum2000global, belkin2001laplacian, van2008visualizing, mcinnes2018umap}.
Natural science researchers often use dimensionality reduction methods for exploratory data analysis or even to focus further experiments
\citep{becht2019dimensionality, kobak2019art, karlov2019chemical, andronov2021exploring, szubert2019structure}.
The main problem with these methods is inevitable distortions \citep{chari2021specious, batson2021topological, wang2021understanding} and incoherent results for different hyperparameters.
These distortions can largely affect global representation structure such as inter-cluster relationships and pairwise distances. As the interpretation of these quantities in some domain such as physics or biology can lead to incorrect conclusions, it is of high importance to preserve them as much as possible.
%Practitioners expect to draw faithful conclusion for visualizations. However, images vary for  different hyperparameters. 
UMAP and t-SNE visualizations are frequently sporadic and cannot be considered as ``canonical'' representation of high-dimensional data. 
An often overlooked issue is the initialization which significantly contributes to the performance of dimensionality reduction methods \citep{kobak2021initialization, wang2021understanding}.
\cite{damrich2021umap} revealed that the UMAP's true loss function is different from the purported from its theory because of negative sampling.
%There is a long-standing question of accessing the quality of low-dimensional embeddings since they are prone to distortions \cite{coenen2019understanding, kobak2019art, wang2021understanding, damrich2021umap, chari2021specious}.
%As a result, UMAP tries to reproduce similarities that only encode the $k$ nearest neighbor graph and fails to preserve global geometry of a manifold.
There is a number of works that try to tackle the distortion problem and preserve as much inter-data relationships as possible. Authors of PHATE \citep{moon2019visualizing} and ivis \citep{szubert2019structure} claim that their methods are able to capture local as well as global features, but provide no theoretical guarantees for this. 
\citep{wagner2021improving} {propose DIPOLE, an approach to dimensionality reduction combining techniques of metric geometry and distributed persistent homology.}
% check this
%\cite{agrawal2021minimum} introduced \texttt{pyMDE}, a computationally efficient GPU-based framework for minimum-distortion embeddings. 

%The framework provides in particular various global and local dimensionality reduction methods.

%Different hyperparameters results on different visualization and lead to different conclusions from the data. 

From a broader view, deep representation learning is also dedicated to obtaining low-dimensional representation of data. Autoencoder \citep{hinton2006reducing} and Variational Autoencoder \citep{kingma2013auto} are mostly used to learn representations of objects useful for solving downstream tasks or data generation. They are not designed for data visualization and fail to preserve simultaneously local and global structure on 2D/3D spaces. Though, their parametric nature makes them scalable and applicable to large datasets, which is why they are used in methods such as parametric UMAP \citep{sainburg2021parametric} and ivis \citep{szubert2019structure} and ours.

%The work \cite{moor2020topological} is the most similar to ours. 
\cite{moor2020topological} proposed TopoAE, including an additional loss for the autoencoder to preserve topological structures of the input space in latent representations. The topological similarity is achieved by retaining similarity in the multi-scale connectivity information.
Our approach has a stronger theoretical foundation and outperforms TopoAE in computational experiments.

An approach for differentiation of persistent homology-based functions was proposed by \cite{carriere2021optimizing}.
\cite{leygonie2021framework} systematizes different approaches to regularisation of persistence diagrams function and defines notions of differentiability for maps to and from the space of persistence barcodes. \cite{luo2021topology} proposed a topology-preserving dimensionality reduction method based on graph autoencoder. \cite{kim2020pllay} proposed a differentiable topological layer for general deep learning models based on persistence landscapes.

%?application to Topological layers 

%UMAP: \cite{mcinnes2018umap}\\
%biologists criticize UMAP \cite{chari2021specious}\\
%t-SNE \cite{van2008visualizing}
%Kobak: The art of using t-SNE for single-cell transcriptomics \cite{kobak2019art}\\
%Laplacian Eigenmaps \cite{belkin2001laplacian}\\
%D. Kobak and G. C. Linderman. Initialization is critical for preserving global data structure in both t-SNE and UMAP. Nature Biotechnology, pages 1–2, 2021. \cite{kobak2021initialization}\\
%ISOMAP: A Global Geometric Framework for Nonlinear Dimensionality Reduction \cite{tenenbaum2000global}\\
%Understanding How Dimension Reduction Tools Work: An Empirical Approach to Deciphering t-SNE, UMAP, TriMap, and PaCMAP for Data Visualization \cite{wang2021understanding} \\
%On UMAP's true loss function \cite{damrich2021umap}\\
%Minimum-Distortion Embeddings \cite{agrawal2021minimum}\\
%Topological AE \cite{moor2020topological}\\

%Chemical space exploration guided by deep neural networks, \cite{karlov2019chemical}\\

%Optimizing both local and global structure without kNN
%No established metrics for ``global structure preserving''

\section{Preliminaries}

\subsection{Topological data analysis, persistent homology}

Topology is often considered to describe the ``shape of data'', that is, multi-scale properties of the datasets. Topological information was generally recognized to be important for various data analysis problems.
In the perspective of the commonly assumed manifold hypothesis \citep{goodfellow2016deep}, datasets are concentrated near low-dimensional manifolds located in high-dimensional ambient spaces. The standard direction is to study topological features of the underlying manifold. The common approach is to cover the manifold via simplices. Given the threshold $\alpha$, we take sets of the points from the dataset $X$ which are pairwise closer than $\alpha$.
The family of such sets is called the Vietoris-Rips simplicial complex.
For further convenience, we introduce the fully-connected weighted graph $\mathcal{G}$ whose vertices are the points from $X$ and whose edges have weights given by the distances between the points. Then, the Vietoris-Rips simplicial complex is defined as:
$$
   \text{VR}_\alpha(\mathcal{G})=\left\{\{{i_0},\ldots,{i_k}\}, i_m \in \text{Vert}(\mathcal{G}) \; \vert \; m_{i,j} \leq \alpha \right\},
$$
where $m_{i,j}$ is the distance between points,  $\text{Vert}(\mathcal{G}) = \{1, \ldots, |X|\}$ is the vertices set of the graph $\mathcal{G}$.
%We note here that the Vietoris-Rips simplicial complex can be defined more abstractly, treating $\{{i_0},\ldots,{i_k}\}$ as vertices of a fully-connected weighted graph and $d(x_{i_m}, x_{j_n})$ - weights associated with the edges. 

For each $\text{VR}_{\alpha}(\mathcal{G})$, we define the vector space $C_k$, which consists of formal linear combinations of all $k$-dimensional simplices from $\text{VR}_{\alpha}(\mathcal{G})$ with modulo 2 arithmetic. The boundary operator $\partial_k: C_k \to C_{k-1}$ maps every simplex to the sum of its facets. One can show that $\partial_k \circ \partial_{k-1}=0$ and the chain complex can be created:
$$
\ldots\rightarrow C_{k+1}\stackrel{\partial_{k+1}}{\rightarrow}C_{k}\stackrel{\partial_{k}}{\rightarrow}C_{k-1}\rightarrow\ldots.
$$
The quotient vector space $H_k=ker(\partial_k)/im(\partial_{k+1})$ is called the $k$-th homology group, elements of $H_k$ are called homology classes. The dimension $\beta_k = dim(H_k)$ is called the $k$-th Betti number and it approximates the number of basic topological features of the manifold represented by the point cloud $X$. 

The immediate problem here is the selection of appropriate $\alpha$ which is not known beforehand. The standard solution is to analyze all $\alpha>0$. Obviously, if $\alpha_1 \le \alpha_2 \le \ldots \le \alpha_m$, then 
$\text{VR}_{\alpha_1}(\mathcal{G}) \subseteq \text{VR}_{\alpha_2}(\mathcal{G}) \subseteq \ldots \subseteq \text{VR}_{\alpha_m}(\mathcal{G})$; the nested sequence is called the filtration.
%Moreover, for each homology class one can track precisely the moments of ``birth'' and ``death'', that is, $\alpha_1$ and $\alpha_2$ when the particular homological class appears for the first time and vanishes. So,
%$\alpha_2-\alpha_1$ describes the ``lifespan'' or persistence of the homology class.
The evolution of cycles across the nested family of simplicial complexes $S_{\alpha_i}$  is canonically decomposed into ``birth'' and ``death'' of  basic topological features,
%across the nested simplicial complexes $S_{\alpha}$
so that a basic feature $c$ appears in $H_k(S_{\alpha})$ at a specific threshold $\alpha_c$ and disappears at a specific threshold $\beta_c$, $\beta_c-\alpha_c$ describes the ``lifespan'' or persistence of the homology class.
The set of the corresponding intervals $[\alpha_c, \beta_c]$ for  the basic homology classes from $H_k$ is called the \textit{persistence barcode}; the whole theory is dubbed the \textit{persistent homology} \citep{chazal2017introduction,Barannikov1994,zomorodian2001computing}.
%; 

\subsection{Representation Topology Divergence (RTD)}
\label{sec:rtd}

The classic persistent homology is dedicated to the analysis of a single point cloud $X$. 
Recently, Representation Topology Divergence (RTD) \citep{barannikov2021representation} was proposed to measure the dissimilarity in the multi-scale topology between two point clouds $X, \tilde{X}$ of equal size $N$ with a one-to-one correspondence between clouds. Let $\mathcal{G}^{w}$, $\mathcal{G}^{\tilde{w}}$ be graphs with weights on edges equal to pairwise distances of $X, \tilde{X}$. 
To provide the comparison, the auxiliary graph $\hat{\mathcal{G}}^{w,\tilde{w}}$ with doubled set of vertices and edge weights matrix $m(w,\tilde{w})$, see details in Appendix \ref{app:rtd_formal}, is created. The persistence barcode of the graph $\hat{\mathcal{G}}^{w,\tilde{w}}$ is called the \textit{R-Cross-Barcode} and it tracks the differences in the multi-scale topology of the two point clouds by comparing their $\alpha$-neighborhood graphs for all $\alpha$. 
%The topological features with longer lifespans indicate in general the essential features.

\begin{figure*}[tbph]
    \centering
    \includegraphics[width=1.0\linewidth]{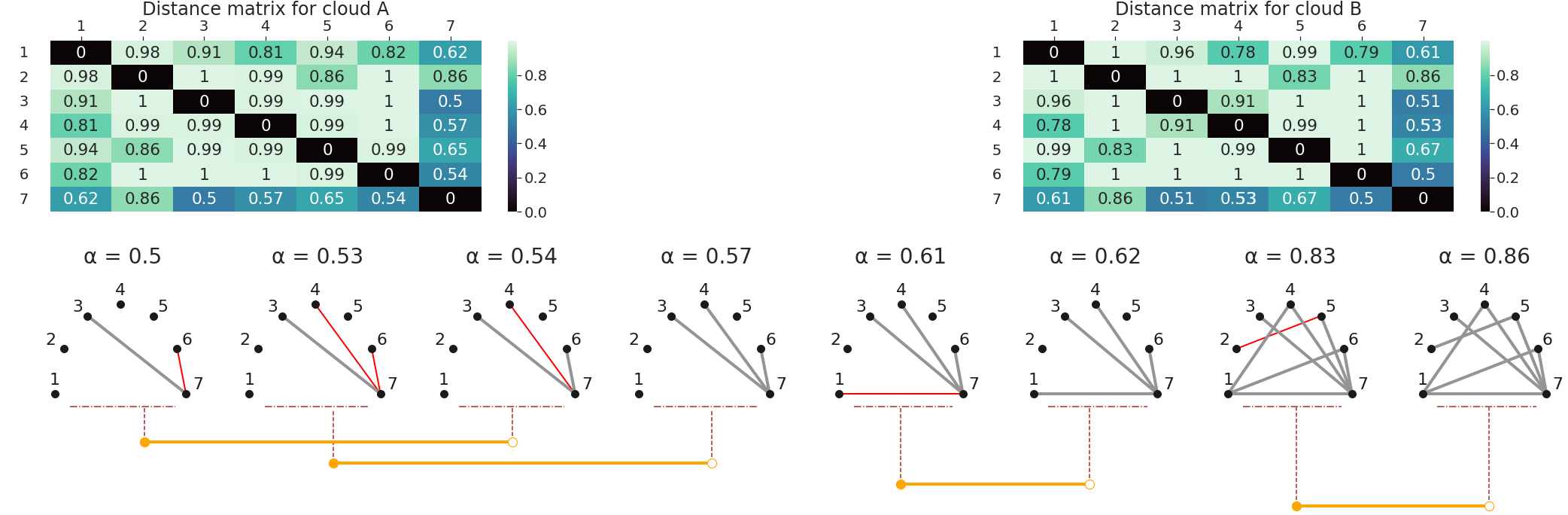}
    \caption{A graphical representation of an R-Cross-Barcode$_1(A, B)$ for the point clouds $A$ and $B$. 
 The pairwise distance matrices for $A$ and $B$ are shown in the top raw. Edges present in the $\alpha$-neighborhood graphs for $B$ but not for $A$ are colored in red. Edges present in the $\alpha$-neighborhood graph for $A$ are colored in grey. The timeline for appearance-disappearance of topological features distinguishing the two graphs is shown. The appearance-disappearance process is illustrated by the underlying bars, connecting the corresponding thresholds.}
    \label{fig:rtd_clarification}
\end{figure*}

Here we give a simple example of an {R-Cross-Barcode}, see also \citep{cherniavskii2022acceptability}. Suppose we have two point clouds $A$ and $B$, of seven points each, with distances between points as shown in the top row of Figure \ref{fig:rtd_clarification}. Consider the R-Cross-Barcode$_1$(A, B), it consists of 4 intervals (the bottom row of the figure). The 4 intervals describe the topological discrepancies between connected components of $\alpha$-neighborhood graphs of $A$ and $B$.

%Let's look at the intervals right-to-left, ``against the passage of time''.
An interval is opened, i.e. a topological discrepancy appears, at threshold $\alpha=\tilde{w}^B_{uv}$  when in the union of $\alpha$-neighborhood graph of $A$ and $B$, two vertex sets $C_1$ and  $C_2$  disjoint at smaller thresholds, are joined into one connected component by the edge $(uv)$ from $B$. 
This interval is closed at  threshold $\alpha=w^A_{u'v'}$ when the two vertex sets $C_1$ and $C_2$ are joined into one connected component in the $\alpha$-neighborhood graph of $A$.
%This interval is opened when in the $\alpha$-neighborhood graph for $B$ appears initially an edge between a vertex from $C_1$ and a vertex from $C_2$.
%(interval is always opened before it is closed because edge $(u, v)$ in $B$ satisfies this condition).

For example, a discrepancy appears at the threshold $\alpha=0.53$ when the vertex sets $\lbrace 4\rbrace$ and $\lbrace 3, 6, 7\rbrace$ are joined into one connected component in the union of neighborhood graphs of $A$ and $B$ by the edge $(4,7)$.
We identify the “death” of this R-Cross-Barcode feature at $\alpha=0.57$, when these two sets are joined into one connected component in the neighborhood graph of cloud A (via the edge $(4, 7)$ in Figure \ref{fig:rtd_clarification} becoming grey). 
%Then we observe its “birth”, when edge between $4$ and $7$ first becomes present in  neighbourhood  graph of $B$ at $\alpha=0.53$ (the corresponding edge now is green on the  Figure \ref{fig:rtd_clarification}).

By definition, $\text{RTD}_k(X, \tilde{X})$ is the sum of intervals' lengths in the \textit{R-Cross-Barcode}$_k (X, \tilde{X})$ and measures its closeness to an empty set.

\begin{proposition}[\cite{barannikov2021representation}]
\label{prop_rtd}
If $RT\!D_k(X, \tilde{X}) = RT\!D_k(\tilde{X}, X)=0$ for all $k \geq 1$, then the barcodes of the weighted graphs $\mathcal{G}^w$ and $\mathcal{G}^{\tilde{w}}$ are the same in any degree. Moreover, in this case the topological features are located in the same places: 
%in $\mathcal{G}^w$ and $\mathcal{G}^{\tilde{w}}$, 
the inclusions $\text{VR}_\alpha(\mathcal{G}^w)\subseteq \text{VR}_\alpha(\mathcal{G}^{\min(w,\tilde{w})})$,  $\text{VR}_\alpha(\mathcal{G}^{\tilde{w}}) \subseteq \text{VR}_\alpha(\mathcal{G}^{\min(w,\tilde{w})})$ induce homology isomorphisms for any threshold~$\alpha$.
\end{proposition}

The Proposition \ref{prop_rtd} is a strong basis for topology comparison and optimization. 
Given a fixed data representation $X$, how to find $\tilde{X}$ lying in a different space, and having a topology similar to $X$, in particular, similar persistence barcodes? Proposition \ref{prop_rtd} states that it is sufficient to minimize $\sum_{i\ge1} \left( \text{RTD}_i(X, \tilde{X})+\text{RTD}_i(\tilde{X},X) \right)$. In most of our experiments we minimized  $\text{RTD}_1(X, \tilde{X})+\text{RTD}_1(\tilde{X},X)$. $\text{RTD}_1$ can be calculated faster than $\text{RTD}_{2+}$, also $\text{RTD}_{2+}$ are often close to zero.
To simplify notation, we denote $\text{RTD}(X,\tilde{X}) := \nicefrac{1}{2}(\text{RTD}_1(X, \tilde{X})+\text{RTD}_1(\tilde{X}, X))$.

\textbf{Comparison with TopoAE loss}. TopoAE \citep{moor2020topological} is the state-of-the-art algorithm for topology-preserving dimensionality reduction. The TopoAE topological loss is based on comparison of minimum spanning trees in $X$ and $\tilde{X}$ spaces. However, it has several weak spots. 
First, when the TopoAE loss is zero there is no guarantee that persistence barcodes of $X$ and $\tilde{X}$ coincide. Second, the TopoAE loss can be discontinuous in rather standard situations, see \mbox{Appendix \ref{app:topoae_critique}}. 
At the same time, RTD loss is continuous, and its nullity guarantees the coincidence of persistence barcodes of $X$ and $\tilde{X}$. \mbox{The continuity of the RTD loss follows from  the stability of the $\text{R-Cross-Barcode}_k$  (Proposition \ref{prop:stab}).}
%, see  \mbox{Appendix \ref{sec:stability}}} for the proofs.
\begin{proposition}\label{prop:stab}
  (a) For any quadruple of edge weights sets $w_{ij}$, ${\tilde{w}}_{ij}$, $v_{ij}$, 
 ${\tilde{v}}_{ij}$ on $\mathcal{G}$:  
$$d_B(\text{R-Cross-Barcode} _ k(w,\tilde{w}), \text{R-Cross-Barcode} _ k(v,\tilde{v})) \leq \max(\max _ {ij} \lvert v _ {ij}-w _ {ij}\rvert ,\max _ {ij}\lvert \tilde{v} _ {ij}-\tilde{w} _ {ij}\rvert).$$  
 \vskip-0.15in
(b) For any  pair of edge weights sets ${w}_{ij}$, $\tilde{w}_{ij}$ on $\mathcal{G}$: 
 \vskip-0.1in
 $$\lVert \text{R-Cross-Barcode}_k(w,\tilde{w})\rVert_B \leq \max_{ij} \lvert w_{ij}-\tilde{w}_{ij}\rvert.$$   
 \vskip-0.1in
(c) The expectation for the bottleneck distance between $\text{R-Cross-Barcode}_k(w,\tilde{w})$ and  $\text{R-Cross-Barcode}_k(w',\tilde{w})$, 
%comparing two pairs of weighted graphs 
%with the edge weights 
where $w_{ij}=w(x_i,x_j)$, $w'_{ij}=w'(x_i,x_j)$, $\tilde{w}_{ij}=\tilde{w}(x_i ,x_j)$,  $w,w',\tilde{w}$ is a triple of metrics on a measure space $(\mathcal{X},\mu)$,  
and  $X=\{x_1,\ldots, x_n\}$, $x_i\in\mathcal{X}$ is a sample from $(\mathcal{X},\mu)$,  is upper bounded by Gromov-Wasserstein distance between $w$ and ${w}'$:  
    $$\int_{\mathcal{X}\times\ldots\times\mathcal{X}}d_B(\text{R-Cross-Barcode}_k(w,\tilde{w}),\text{R-Cross-Barcode}_k(w',\tilde{w}))  d\mu^{\otimes n }\leq n\,GW(w,{w}').$$   
(d) The expectation for the bottleneck norm of $\text{R-Cross-Barcode} _ k(w,\tilde{w})$ for two weighted graphs with  edge weights  $w _ {ij}=w(x _ i,x _ j)$, $\tilde{w} _ {ij}=\tilde{w}(x _ i,x _ j )$, where $w ,\tilde{w}$ is a pair of metrics on a measure space $(\mathcal{X},\mu)$, and    $X  =\{x _ 1,\ldots, x _ n\}$, $x _ i\in\mathcal{X}$ is a sample from $(\mathcal{X},\mu)$,
     is upper bounded by Gromov-Wasserstein distance between $w$ and $\tilde{w}$: 
\vskip-0.1in
$$   \int_{\mathcal{X}\times\ldots\times\mathcal{X}}\lVert \text{R-Cross-Barcode}_k(w,\tilde{w})\rVert_B  d\mu^{\otimes n }\leq n\,GW(w,\tilde{w}).$$   
\end{proposition}
 \vskip-0.1in
The proofs are given in  \mbox{Appendix \ref{sec:stability}}.

\section{Method}
\vspace{-0.1in}
\subsection{Differentiation of RTD}
\vspace{-0.03in}
\begin{wrapfigure}{r}{0.35\textwidth}
%\begin{figure}[t]
    \centering
    %\vskip-0.75in
    \vskip-0.5in
    \includegraphics[width=0.35\textwidth]{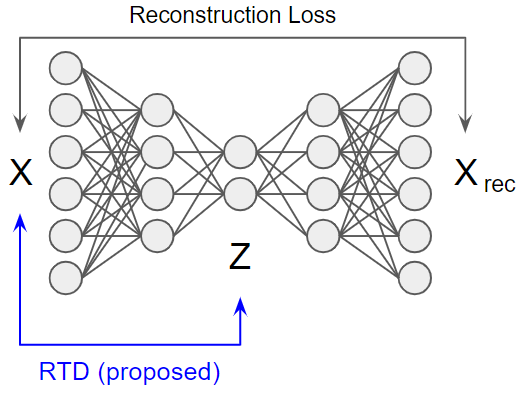}
    \caption{RTD Autoencoder}
    \label{fig:rtd_autoencoder}
    \vskip-0.2in
%\end{figure}
\end{wrapfigure}

We propose to use RTD as a loss in neural networks. Here we describe our approach to RTD differentiation. 
Denote by $\Sigma_k$ the set of all $k-$simplices in the Vietoris-Rips complex of the graph $\hat{\mathcal{G}}^{w,\tilde{w}}$, and by $\mathcal{T}_k$ the set of all intervals in the $\textit{R-Cross-Barcode}_k(X, \tilde{X})$. Fix (an arbitrary) strict order on $\mathcal{T}_k$.

There exists a function $ f_k: ~ \cup_{(b_i, d_i) \in \mathcal{T}_k} \lbrace b_i, d_i \rbrace\rightarrow \Sigma_k$ that maps $b_i$ (or $d_i$) to a simplex $\sigma$ whose appearance leads to ``birth'' (or ``death'') of the corresponding homological class.
%Here $b_i$ and $d_i$ are the filtration values at which simplices $f_k(b_i)$ and $f_k(d_i)$ join the filtration. 
%They depend on weights of graph edges as
%\begin{equation*}
%b_i = g_\sigma = \max_{i, j \in \sigma} %m_{i, j}
%\end{equation*}
Let 
\begin{equation*}
m_\sigma = \max_{i, j \in \sigma} m_{i, j}
\end{equation*} denote the function of $m_{ij}$ equal to the filtration value at which the simplex $\sigma$ joins the filtration.
Since $\frac{\partial ~ \text{RTD}_k(X, \tilde{X})}{\partial d_i}=-\frac{\partial ~ \text{RTD}_k(X, \tilde{X})}{\partial b_i}=1$, we obtain the following equation for the subgradient
\begin{equation*}
    \frac{\partial ~ \text{RTD}_k(X, \tilde{X})}{\partial m_\sigma} = \sum_{i \in \mathcal{T}_k} \mathbb{I}\lbrace f_k(d_i) = \sigma \rbrace - \sum_{i \in \mathcal{T}_k} \mathbb{I}\lbrace f_k(b_i) = \sigma \rbrace.
\end{equation*}
Here, for any $\sigma$ no more than one term has non-zero indicator.
   Then 
 \begin{equation*}
     \frac{\partial ~ \text{RTD}_k(X, \tilde{X})}{\partial m_{i, j}} = \sum_{\sigma \in \Sigma_k}\frac{\partial ~ \text{RTD}_k(X, \tilde{X})}{\partial m_\sigma}  \frac{\partial m_\sigma}{\partial m_{i, j}}
 \end{equation*}
%Function $g_\sigma$ are differentiable \citep{leygonie2021framework}.
%and so is $f_k \circ g_k$.
%And so, we obtain the subgradient
%$$
%\frac{\partial ~ \text{RTD}(X, \hat{X})}{\partial m_{i, j}} = \sum_{\sigma \in \Sigma}\frac{\partial ~ \text{RTD}(X, \hat{X})}{\partial g_\sigma}  \frac{\partial g_\sigma}{\partial m_{i, j}}.
%$$
The only thing that is left is to obtain subgradients of $\text{RTD}(X, \tilde{X})$ by points from $X$ and $\tilde{X}$ .
Consider (an arbitrary) element $m_{i, j}$ of matrix $m$. There are 4 possible scenarios:
\vspace{-0.5pc}
\begin{enumerate}
    \item $i, j \leq N$, in other words $m_{i, j}$ is from the upper-left quadrant of $m$. Its length is constant and thus $\forall l: \frac{\partial m_{i, j}}{\partial X_l} =  \frac{\partial m_{i, j}}{\partial \tilde{X_l}} = 0$.
    \item $i \leq N < j$, in other words $m_{i, j}$ is from the upper-right quadrant of $m$. Its length is computed as Euclidean distance and thus $\frac{\partial m_{i, j}}{\partial X_i} = \frac{X_i - X_{j - N}}{\lvert\rvert X_i - X_{j - N}\lvert\rvert_2}$ (similar for $X_{N - j}$).
    \item $j \leq N < i$, similar to the previous case.
    \item $N < i, j $, in other words $m_{i, j}$ is from the bottom-right quadrant of $m$. Here we have subgradients like
    $$\frac{\partial m_{i, j}}{\partial X_{i - N}} =  \frac{X_i - X_{j - N}}{\lvert\rvert X_i - X_{j - N}\lvert\rvert_2}\mathbb{I}\lbrace w_{i - N, j - N} < \tilde{w}_{i - N, j - N}\rbrace $$
    Similar for $X_{j-N}, \tilde{X}_{i-N}$ and $\tilde{X}_{j-N}$.
\end{enumerate}
\vspace{-0.5pc}
Subgradients $\frac{\partial ~ \text{RTD}(X, \tilde{X})}{\partial X_{i}}$ and $\frac{\partial ~ \text{RTD}(X, \tilde{X})}{\partial \tilde{X}_{i}}$ can be derived from the beforementioned using the chain rule and the formula of full (sub)gradient. Now we are able to minimize $\text{RTD}(X, \tilde{X})$ by methods of (sub)gradient optimization. We discuss some possible tricks for improving RTD differentiation in Appendix \ref{app:rtd_tricks}.

\subsection{RTD Autoencoder}

Given the data $X = \{x_i\}_{i=1}^n$, $x_i \in \mathbb{R}^d$, in high-dimensional space, our goal is to find the representation in low-dimensional space $Z = \{ z_i\}$, $z_i \in \mathbb{R}^p$. For the visualization purposes, $p=2,3$.
%Existing methods preserve distances or neighbors between $X$ and $Z$. 
Our idea is to find a representation $Z$ which preserves \textit{persistence barcodes}, that is, multi-scale topological properties of the point clouds, as much as possible. The straightforward approach is to solve
$\min_{Z} \text{RTD}(X, Z),$
where the optimization is performed over $n$ vectors $z_i \in \mathbb{R}^p$, in the flavor similar to UMAP and t-SNE. This approach is workable albeit very time-consuming and could be applied only to small datasets, see Appendix \ref{app:pure_rtd}.
A practical solution is to learn representations via the encoder network $E(w,x): X \to Z$, see Figure \ref{fig:rtd_autoencoder}.
%Let $E(w, x)$ be an encoder, $D(w, z)$ be a decoder.

\textbf{Algorithm}. Initially, we train the autoencoder for $E_1$ epochs with the reconstruction loss \mbox{$\frac{1}{2}||X - X_{rec}||^2$} only. Then, we train for $E_2$ epochs with the loss $\frac{1}{2}||X - X_{rec}||^2 + \text{RTD}(X, Z)$. Both losses are calculated on mini-batches. The two-step procedure speedups training since calculating $\text{RTD}(X, Z)$ for the untrained network takes much time.

%\textbf{Modifications}.
%We emphasize that calculating RTD makes sense only of for quite large batches points since topology of a single point is trivial. 

%Then we follow the algorithm:
%\begin{enumerate}
%    \item Train the autoencoder:
%    $$
%    \min_{w_1, w_2} \sum_{i=1}^n (D(w_2, E(w_1, x_i)) - x_i)^2
%    $$
%    \item Fine-tune the encoder with the RTD loss:
%    $$
%    \min_{w_1} \mathbb{E}_{X_b \sim X} RTD(X_b, E(w_1, X_b)),
%    $$
%    where $X_b$ is a batch - a subset of $b$ objects randomly sampled from $X$. 
%\end{enumerate}

\section{Experiments}
\label{sec:experiments}

% \begin{table}
%     \centering
%     \caption{Datasets description.}
%     \label{tab:dataset_description}
%     \begin{tabular}{lllll}
%     \toprule
%     Dataset name & Total size & Nature & Dimension & Description \\
%     \toprule
%     Circle & $1\times 10^2$ & Synthetic & $2$ & Sampled from a 2D circle \\
%     \midrule
%     Random & $5 \times 10^2$ & Synthetic & $2$ & Sampled from a 2D unit square \\
%     \midrule
%     2 Clusters & $2 \times 10^2$ & Synthetic & $2$ & Two Gaussian 2D clusters, one sparse, one dense \\
%     \midrule
%     3 Clusters & $3 \times 10^2$ & Synthetic & $2$ & Three Gaussian 2D clusters, two are located much \\
%     & & & & closer to each other than the remaining one \\
%     \midrule
%     Mammoth & $50 \times 10^3$ & Real & $3$ & The skeleton of a mammoth \\
%     \midrule
%     F-MNIST & $70 \times 10^3$ & Real & $784$ & Each example is a $28 \times 28$ grayscale image, \\
%     & & & & depicting clothes, associated with a label from 10 classes. \\
%     \midrule
%     COIL-20 & $1440$ & Real & $16384$ & Each example is a $128 \times 128$  greyscale image,\\
%     & & & & associated with a label from 20 classes, each designating an object\\
%     & & & & under 72 different rotations spanning 360 degrees. \\
%     \midrule
%     Calorimeter & $1.5 \times 10^4$ & Real & $507$ & A collection of 3-layer calorimeter showers \\
%     & & & & corresponding to one particle (positrons, photons, \\
%     & & & & or charged pion).  \\
%     \midrule
%     C.elegans & $86024$ & Real & $100$ & \\
%     \bottomrule
%     \end{tabular}
% \end{table}

\begin{figure}[b!]
    \centering
    \includegraphics[width=\textwidth]{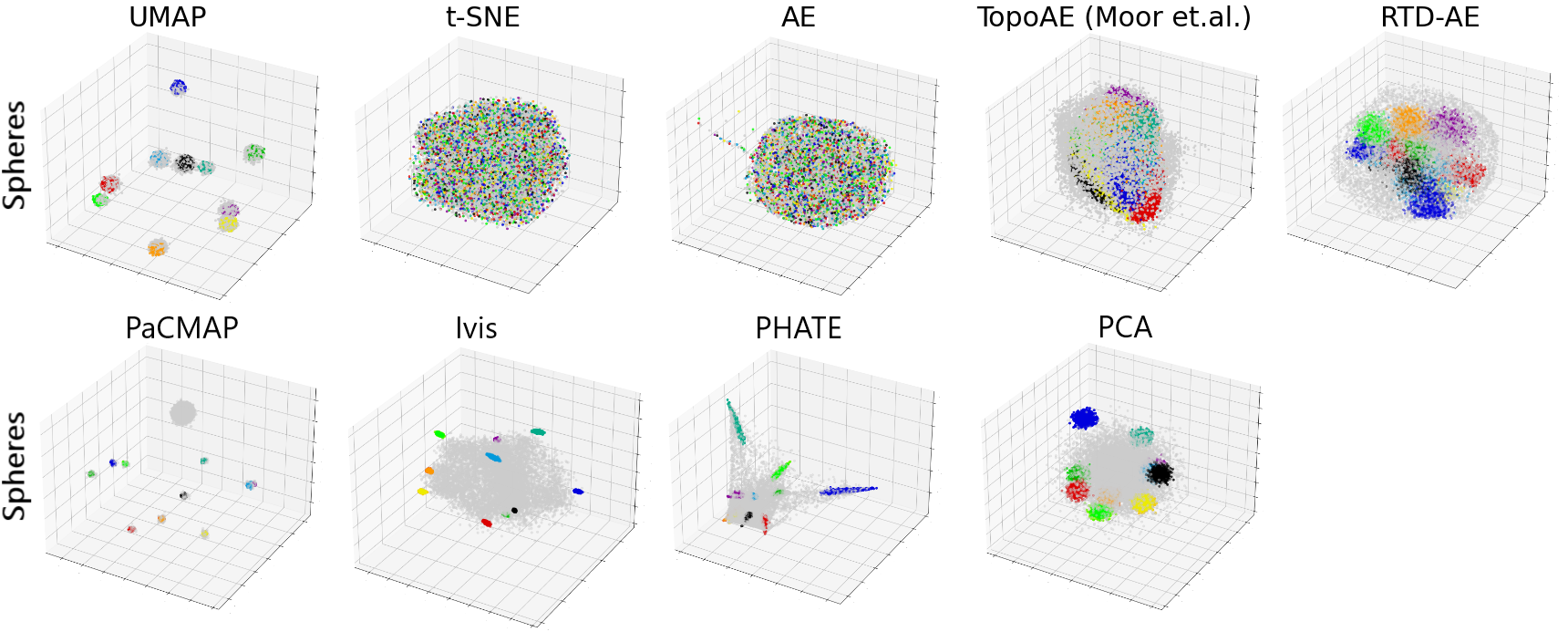}
    \caption{Results on dimensionality reduction to 3D-space}
    \label{fig:synthetic3d}
\end{figure}

\begin{table}[b!]
  \caption{Quality of data manifold global structure preservation at projection from 101D into 3D space.}
  \label{tbl:measure_synthetic3d}
  \centering
  \begin{tabular}{llllll}
    \toprule
    & & \multicolumn{4}{c}{Quality measure}           \\
    \cmidrule(r){3-6}
    Dataset & Method   & L. C. & W. D. $H_{0}$ & T. A. & RTD \\
    \midrule
    Spheres 3D & t-SNE & {0.087} & 47.89 $\pm$ 2.59 & 0.206 $\pm$ 0.01 & 37.32 $\pm$ 1.44 \\
            & UMAP & 0.049 & 48.31 $\pm$ 1.83 & 0.313 $\pm$ 0.03 & 44.70 $\pm$ 1.47 \\
            
            & PaCMAP & 0.394 & 46.48 $\pm$  1.61 & 0.156 $\pm$ 0.02 & 45.88 $\pm$ 1.51  \\
            & PHATE & 0.302 & 48.78 $\pm$ 1.65 & 0.207 $\pm$ 0.02 & 44.05 $\pm$  1.42 \\
            & PCA & 0.155 & 47.15 $\pm$ 1.89 & 0.174 $\pm$ 0.02 & 38.96 $\pm$  1.25 \\
            & MDS & N.A. & N.A. & N.A. & N.A.\\
            & Ivis & 0.257 & 46.32 $\pm$ 2.04 & 0.130 $\pm$ 0.01 & 41.15 $\pm$ 1.28 \\
            
            & AE & 0.441 & \textbf{45.07 $\pm$ 2.27} & 0.333 $\pm$ 0.02 & 39.64 $\pm$ 1.45 \\
            & TopoAE & {0.424} & 45.89 $\pm$ 2.35 & 0.274 $\pm$ 0.02 & 38.49 $\pm$ 1.59 \\
            & RTD-AE & \textbf{0.633} & \textbf{45.02 $\pm$ 2.69} & \textbf{0.346 $\pm$ 0.02} & 
            \textbf{35.80 $\pm$ 1.63} \\
    \bottomrule
  \end{tabular}
\end{table}

% W.D. H1
% & 1.326 $\pm$ 0.12 
% &  0.192 $\pm$ 0.01 
% & 0.347 $\pm$ 0.03 
% & 0.224 $\pm$ 0.04 
% & 0.768 $\pm$ 0.08 
% & 0.580 $\pm$ 0.07 
% & 0.582 $\pm$ 0.07 
% & 1.411 $\pm$ 0.10 
% & 1.060 $\pm$ 0.10 

In computational experiments, we perform dimensionality reduction to high-dimensional and 2D/3D space for ease of visualization.
We compare original data with latent representations by (1) linear correlation of pairwise distances, (2) Wasserstein distance (W.D.) between $H_{0}$ persistence barcodes \citep{chazal2017introduction}, (3) triplet distance ranking accuracy \citep{wang2021understanding} (4) RTD.
All of the quality measures are tailored to evaluate how the manifold's global structure and topology are preserved. 
We note that RTD, as a quality measure, provides a more precise comparison of topology than the W.D. between $H_{0}$ persistence barcodes. First, RTD takes into account the localization of topological features, while W.D. does not. Second, W.D. is invariant to permutations of points, but we are interested in comparison between original data and latent representation where natural one-to-one correspondence holds.

We compare the proposed RTD-AE with t-SNE \citep{van2008visualizing}, UMAP \citep{mcinnes2018umap}, TopoAE \citep{moor2020topological}, vanilla autoencoder (AE), PHATE \citep{moon2019visualizing}, Ivis \citep{szubert2019ivis}, PacMAP \citep{wang2021understanding}. 
%Table \ref{tbl:measure}  presents the results.
%In order to illustrate the issues of representations produced by other methods, we generated a number of synthetic datasets.
The complete description of all the used datasets can be found in Appendix \ref{app:datasets}.
See hyperparameters in Appendix \ref{app:hyperparams}.

%
% REAL-WORLD DATA RESULTS
%
\begin{table}[t]
\caption{Quality of data manifold global structure preservation at projection into 16D space.}
\centering
\label{tbl:real16dim}
\begin{tabular}{llllll}
    \toprule
    & & \multicolumn{4}{c}{Quality measure}           \\
    \cmidrule(r){3-6}
    Dataset & Method   & L. C. & W. D. $H_{0}$ &  T. A. & RTD \\
    \midrule
    F-MNIST  & UMAP & 0.602 & 592.0 $\pm$ 3.9 & 0.741 $\pm$ 0.018 & 12.31 $\pm$ 0.44\\
             & PaCMAP & 0.600 & 585.9 $\pm$ 3.2 & 0.741 $\pm$ 0.013 & 12.72 $\pm$ 0.48 \\
             & Ivis & 0.582 & 552.6 $\pm$ 3.5 & 0.718 $\pm$ 0.014 & 10.76 $\pm$ 0.30 \\
             & PHATE & 0.603 & 576.4 $\pm$ 4.4 & 0.756 $\pm$ 0.016 & 10.72 $\pm$ 0.15\\
             & AE & 0.879 & 320.5 $\pm$ 1.9 & 0.850 $\pm$ 0.004 & 5.52 $\pm$ 0.17  \\
             & TopoAE & 0.905 & 190.7 $\pm$ 1.2 & 0.867 $\pm$ 0.006 & 3.69 $\pm$ 0.24 \\
             & RTD-AE & \textbf{0.960} & \textbf{181.2 $\pm$ 0.8} & \textbf{0.907 $\pm$ 0.004} & \textbf{3.01 $\pm$ 0.13}\\
    \midrule
   MNIST & UMAP & 0.427 & 879.1 $\pm$ 5.6 & 0.625 $\pm$ 0.016 & 17.62 $\pm$ 0.73 \\
        & PaCMAP & 0.410 & 887.5 $\pm$ 6.1 & 0.644 $\pm$ 0.012 & 20.07 $\pm$ 0.70 \\
        & Ivis & 0.423 & 712.6 $\pm$ 5.0 & 0.668 $\pm$ 0.013 & 12.40 $\pm$ 0.32 \\
        & PHATE & 0.358 & 819.5 $\pm$ 4.0 & 0.626 $\pm$ 0.018 & 15.01 $\pm$ 0.25\\
        & AE   & 0.773 & 391.0 $\pm$ 2.9 & 0.771 $\pm$ 0.010 & 7.22 $\pm$ 0.14\\
        & TopoAE & 0.801 & 367.5 $\pm$ 1.9 & 0.796 $\pm$ 0.014 & 5.84 $\pm$ 0.19 \\
        & RTD-AE & \textbf{0.879} & \textbf{329.6 $\pm$ 2.6} & \textbf{0.833 $\pm$ 0.006} & \textbf{4.15 $\pm$ 0.18} \\
   \midrule
   COIL-20 & UMAP & 0.301 & 274.7 $\pm$ 0.0 &  0.574 $\pm$ 0.011 & 15.99 $\pm$ 0.52\\
            & PaCMAP & 0.230 & 273.5 $\pm$ 0.0 & 0.548 $\pm$ 0.012 & 15.18 $\pm$ 0.35 \\
            & Ivis & N.A. & N.A. & N.A. & N.A. \\
            & PHATE & 0.396 & 250.7 $\pm$ 0.000 & 0.575 $\pm$ 0.014 & 13.76 $\pm$ 0.78\\
            & AE   & 0.834 & 183.6 $\pm$ 0.0 &  0.809 $\pm$ 0.008 & 8.35 $\pm$ 0.15\\
            & TopoAE & 0.910 & 148.0 $\pm$ 0.0 & 0.822 $\pm$ 0.020 & 6.90 $\pm$ 0.19\\
            & RTD-AE & \textbf{0.944} & \textbf{88.9 $\pm$ 0.0} & \textbf{0.892 $\pm$ 0.007} & \textbf{5.78 $\pm$ 0.10}  \\
   \midrule
   scRNA mice & UMAP & 0.560 & 1141.0 $\pm$ 0.0 & 0.712 $\pm$ 0.010 & 21.30 $\pm$ 0.17\\
            & PaCMAP & 0.496 & 1161.3 $\pm$ 0.0 & 0.674 $\pm$ 0.016 & 21.89 $\pm$ 0.13\\
            & Ivis & 0.401 & 1082.6 $\pm$ 0.0 & 0.636 $\pm$ 0.007 & 22.56 $\pm$ 1.13 \\
            & PHATE & 0.489 & 1134.6 $\pm$ 0.0 & 0.722 $\pm$ 0.013 & 21.34 $\pm$ 0.32 \\
            & AE & 0.710 & 1109.2 $\pm$ 0.0 & 0.788 $\pm$ 0.013 & 20.80 $\pm$ 0.16\\
            & TopoAE & 0.634 & \textbf{826.0 $\pm$ 0.0} & 0.748 $\pm$ 0.010 & \textbf{15.37 $\pm$ 0.22}\\
            & RTD-AE & \textbf{0.777} & 932.9 $\pm$ 0.0 & \textbf{0.802 $\pm$ 0.006} & 17.03 $\pm$ 0.15\\
   \midrule
   scRNA melanoma & UMAP & 0.474 & 1416.9 $\pm$ 9.2 & 0.682 $\pm$ 0.013 & 20.02 $\pm$ 0.35\\
            & PaCMAP & 0.357 & 1441.8 $\pm$ 9.1 & 0.681 $\pm$ 0.014 & 20.53 $\pm$ 0.36\\
            & Ivis & 0.465 & 1168.0 $\pm$ 11.4 & 0.653 $\pm$ 0.016 & 16.31 $\pm$ 0.28 \\
            & PHATE & 0.427 & 1427.5 $\pm$ 9.1 & 0.687 $\pm$ 0.018 & 20.18 $\pm$ 0.41 \\
            & AE & 0.458 & 1345.9 $\pm$ 11.3 & 0.708 $\pm$ 0.016 & 19.50 $\pm$ 0.37\\
            & TopoAE & 0.544 & 973.7 $\pm$ 11.1 & 0.709 $\pm$ 0.011 & 13.41 $\pm$ 0.35\\
            & RTD-AE & \textbf{0.684} & \textbf{769.5 $\pm$ 11.5} & \textbf{0.728 $\pm$ 0.017} & \textbf{10.35 $\pm$ 0.33}\\
   \bottomrule
\end{tabular}
\vskip-0.2in
\end{table}
\vskip-0.3in
\subsection{Synthetic datasets}

We start with the synthetic dataset ``Spheres'': eleven 100D spheres in the 101D space, any two of those do not intersect and one of the spheres contains all other inside.
For the visualization, we perform dimensionality reduction to 3D space.
Figure \ref{fig:synthetic3d} shows the results: RTD-AE is the best one preserving the nestedness for the ``Spheres'' dataset. 
Also, RTD-AE outperforms other methods by quality measures, see 
Table \ref{tbl:measure_synthetic3d}. We were unable to run MDS on ``Spheres'' dataset because it was too large for that method.
See more results in Appendix \ref{app:rtd_3d}. 

% RTD respects the global structure, shown in all of the datasets.
% Doesn't typically suffer from squeezing (or bottleneck) issue, unlike AE
% Respects the clusters density and their relative distances, unlike t-SNE and UMAP

%The dimensionality reduction to 3D space can be more challenging topologically than reduction to 2D space. Our datasets in this experiment are (a) ``Spheres'': 100D spheres in the 101D space, any two of those do not intersect and one of the spheres contains all other inside; (b) ``Torus'': a torus in 100D space (more details are in Appendix \ref{app:rtd_3d}). Figure \ref{fig:synthetic3d} shows the results: RTD-AE is the best one preserving the nestedness for the ``spheres'' dataset. 
%Also, RTD-AE outperforms other methods by quality measures, see 
%Table \ref{tbl:measure_synthetic3d}. We were unable to run MDS on ``Spheres'' dataset because it was too large for that method.

\subsection{Real world datasets}

% Harder to understand as the initial shape is unknown
% Bottleneck issue (Squeezing): Mammoth, F-MNIST, Calorimeter
% Failure to separate classes on COIL-20 and F-MNIST: both AE, TopoAE and RTD

%Regarding the real world datasets, it is harder to analyse visualizations as the true shape of the data is unknown in general. Despite this, we can still evaluate representations based on our rational assumptions about the structure of the data. 
We performed experiments with a number of real-world datasets: MNIST \citep{lecun1998gradient}, F-MNIST \citep{xiao2017online}, COIL-20 \citep{nene1996columbia}, scRNA mice \citep{yuan2017challenges}, scRNA melanoma \citep{tirosh2016dissecting} with latent dimension of 16 and 2, see Tables \ref{tbl:real16dim}, \ref{tbl:measure_real}. The choice of scRNA datasets was motivated by the increased importance of dimensionality reduction methods in natural sciences, as was previously mentioned. RTD-AE is consistently better than competitors; moreover, the gap in metrics for the latent dimension 16 is larger than such for the latent dimension 2 (see Appendix \ref{app:real_world_2d}). \footnote{PHATE execution take too much time and its results are no presented for many datasets.}
For the latent dimension 2, RTD-AE is the first or the second one among the methods by the quality measures (see Table \ref{tbl:measure_real}, Figure \ref{fig:results_real} in Appendix \ref{app:real_world_2d}).
We conclude that the proposed RTD-AE does a good job in preserving global structure of data manifolds.

For the ``Mammoth'' \citep{mammoth1} dataset (Figure \ref{fig:mammoth}) we did dimensionality reduction 3D $\to$ 2D. Besides good quality measures, RTD-AE produced an appealing 2D visualization: both large-scale (shape) and low-scale (chest bones, toes, tusks) features are preserved.

\begin{figure}[t]
    \centering
    \includegraphics[width=\linewidth]{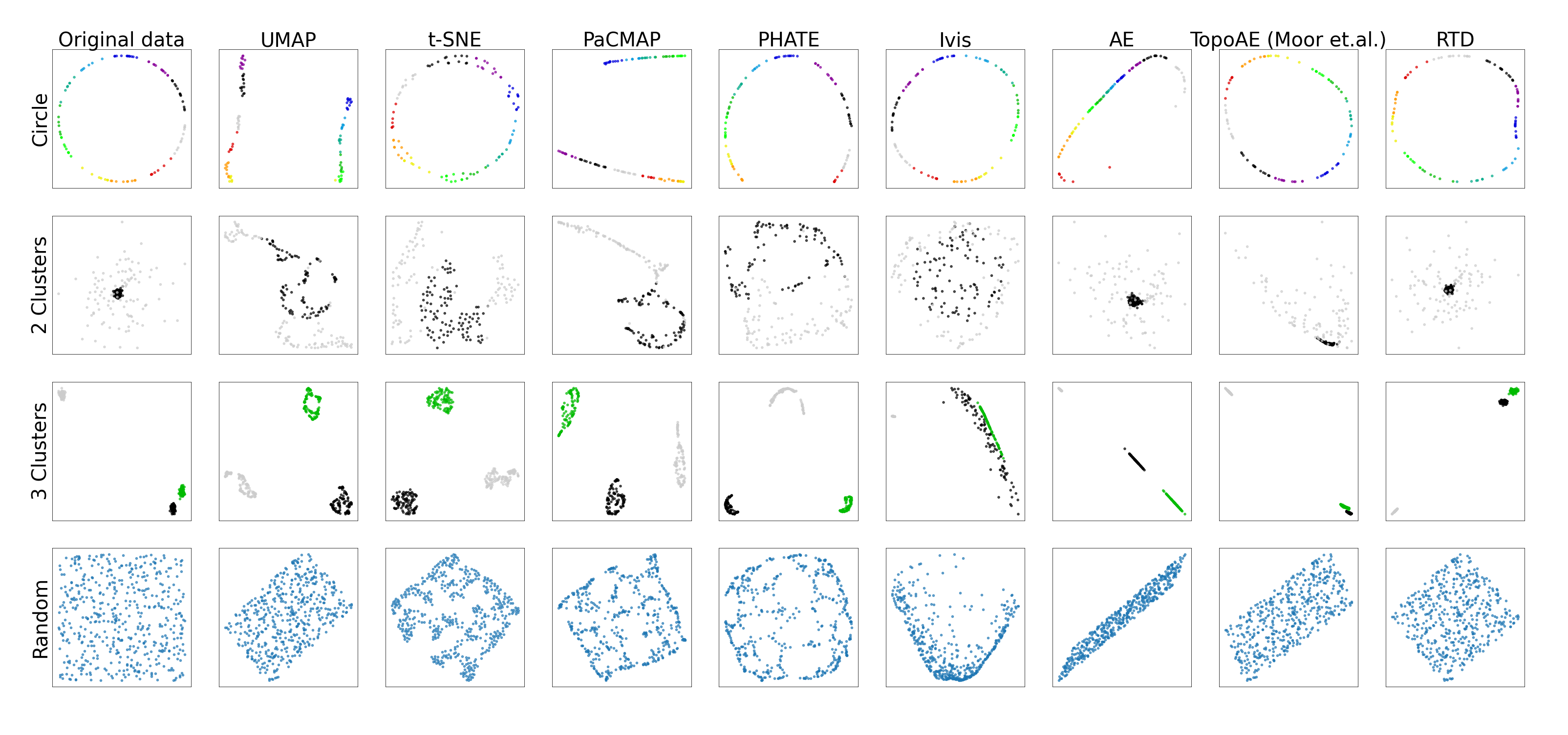}
    \caption{Results on synthetic 2D data. First column: original data. Other columns: results of dimensionality reduction methods.}
    \label{fig:results_synthetic}
\end{figure}

\vskip-0.2in
\subsection{Analysis of distortions}\label{subsec:2d2d}
\vskip-0.1in
Next, to study distortions produced by various dimensionality reduction methods we learn transformation from 2D to 2D space, see Figure \ref{fig:results_synthetic}.
Here, we observe that RTD-AE in general recovers the global structure for all of the datasets. %The produced visualizations are presented in Figure \ref{fig:results_synthetic}. 
RTD-AE typically does not suffer from the squeezing (or bottleneck) issue, unlike AE, which is noticeable in ``Random'', ``3 Clusters'' and ``Circle''. Whereas t-SNE and UMAP struggle to preserve cluster densities and intercluster distances, RTD-AE manages to do that in every case. It does not cluster random points together, like t-SNE. Finally, the overall shape of representations produced by RTD-AE is consistent, it does not tear apart close points, which is something UMAP does in some cases, as shown in the ``Circle'' dataset. The metrics, presented in the Table \ref{tbl:measure_synthetic} in Appendix \ref{app:synthetic_2d}, also confirm the statements above. RTD-AE has typically higher pairwise  distances linear correlation and triplet accuracy, which accounts for good multi-scale properties, while having a lower Wasserstein distance between persistence barcodes.
\vskip-0.1in
\subsection{Limitations and computational complexity}
\label{sec:limitations}
\vskip-0.1in
The main source of complexity is RTD computation. For the batch size $b$, object dimensionality $d$ and latent dimensionality $k$, the complexity is $O(b^{2}(d+k))$ operations since all the pairwise distances should be calculated. The R-Cross-Barcode computation is at worst cubic in the number of simplices involved.
However, the computation is often quite fast for batch sizes $\leq256$ since the boundary matrix is typically sparse for real datasets.
The selection of simplices whose addition leads to ``birth'' or ``death'' of the corresponding homological class doesn't take extra time.
For RTD calculation and differentiation, we used  GPU-optimized software.
As calculation relies heavily on the batch size, the training time of RTD-AE ranges from 1.5x the time of the basic autoencoder at batch size 8 to 4-6x the time in case of batch 512. For COIL-20, the it took $\sim$10 minutes to train a basic AE and $\sim$20 minutes for RTD-AE.
Overall, the computation of a R-Cross-Barcode takes a similar time as in the previous step even on datasets of big dimensionality.
\vskip-0.2in
\subsection{Discussion}
\vskip-0.1in
Experimental results show that RTD-AE better preserves the data manifold global structure than its competitors. The most interesting comparison is with TopoAE, the state-of-the-art,  which uses an alternative topology-preserving loss. The measures of interest for topology comparison are the Wasserstein distances between persistence barcodes.
Tables \ref{tbl:real16dim}, \ref{tbl:measure_synthetic}, \ref{tbl:measure_real} show that RTD-AE is better than TopoAE. RTD minimization has a stronger theoretical foundation than the loss from TopoAE (see Section \ref{sec:rtd}).
\vskip-0.5in
\section{Conclusions}
\vskip-0.13in
In this paper, we have proposed an approach for topology-preserving representation learning (dimensionality reduction). The topological similarity between data points in original and latent spaces is achieved by minimizing the Representation Topology Divergence (RTD) between original data and latent representations. Our approach is theoretically sound: RTD=0 means that persistence barcodes of any degree coincide and the topological features are located in the same places. We proposed how to make RTD differentiable and implemented it as an additional loss to the autoencoder, constructing RTD-autoencoder (RTD-AE).
Computational experiments show that the proposed RTD-AE better preserves the global structure of the data manifold (as measured by linear correlation, triplet distance ranking accuracy, Wasserstein distance between persistence barcodes) than popular methods t-SNE and UMAP. Also, we achieve higher topological similarity than the alternative TopoAE method. Of course, the application of RTD loss is not limited to autoencoders and we expect more deep learning applications involving one-to-one correspondence between points. The main limitation is that calculation of persistence barcodes and RTD, in particular, is computationally demanding. We see here another opportunity for further research.

\section*{Acknowledgements}
The work was supported by the Analytical center under the RF Government (subsidy agreement 000000D730321P5Q0002, Grant No. 70-2021-00145 02.11.2021)

\section*{Reproducibility Statement}
\vskip-0.1in
To provide reproducibility, we release the source code of the proposed RTD-AE, see section \ref{sec:intro}, for hyperparameters see Appendix \ref{app:hyperparams}. For other methods, we used either official implementations 
 or implementations from scikit-learn with default hyperparameters. We used public datasets (see Section \ref{sec:experiments}, Appendix \ref{app:datasets}). We generated several synthetic datasets and made the generating code available.

%\clearpage

\bibliography{references}

\begin{thebibliography}{42}
\providecommand{\natexlab}[1]{#1}
\providecommand{\url}[1]{\texttt{#1}}
\expandafter\ifx\csname urlstyle\endcsname\relax
  \providecommand{\doi}[1]{doi: #1}\else
  \providecommand{\doi}{doi: \begingroup \urlstyle{rm}\Url}\fi

\bibitem[Andronov et~al.(2021)Andronov, Fedorov, and
  Sosnin]{andronov2021exploring}
Mikhail Andronov, Maxim~V Fedorov, and Sergey Sosnin.
\newblock Exploring chemical reaction space with reaction difference
  fingerprints and parametric {t-SNE}.
\newblock \emph{ACS omega}, 6\penalty0 (45):\penalty0 30743--30751, 2021.

\bibitem[Barannikov(1994)]{Barannikov1994}
Serguei Barannikov.
\newblock The framed {M}orse complex and its invariants.
\newblock \emph{Advances in Soviet Mathematics}, 21:\penalty0 93--115, 1994.

\bibitem[Barannikov et~al.(2021)Barannikov, Trofimov, Sotnikov, Trimbach,
  Korotin, Filippov, and Burnaev]{barannikov2021manifold}
Serguei Barannikov, Ilya Trofimov, Grigorii Sotnikov, Ekaterina Trimbach,
  Alexander Korotin, Alexander Filippov, and Evgeny Burnaev.
\newblock Manifold {T}opology {D}ivergence: a framework for comparing data
  manifolds.
\newblock \emph{Advances in Neural Information Processing Systems},
  34:\penalty0 7294--7305, 2021.

\bibitem[Barannikov et~al.(2022)Barannikov, Trofimov, Balabin, and
  Burnaev]{barannikov2021representation}
Serguei Barannikov, Ilya Trofimov, Nikita Balabin, and Evgeny Burnaev.
\newblock {R}epresentation {T}opology {D}ivergence: a method for comparing
  neural network representations.
\newblock In \emph{International Conference on Machine Learning (ICML)}, pp.\
  1607--1626. PMLR, 2022.

\bibitem[Batson et~al.(2021)Batson, Haaf, Kahn, and
  Roberts]{batson2021topological}
Joshua Batson, C~Grace Haaf, Yonatan Kahn, and Daniel~A Roberts.
\newblock Topological obstructions to autoencoding.
\newblock \emph{Journal of High Energy Physics}, 2021\penalty0 (4):\penalty0
  1--43, 2021.

\bibitem[Becht et~al.(2019)Becht, McInnes, Healy, Dutertre, Kwok, Ng, Ginhoux,
  and Newell]{becht2019dimensionality}
Etienne Becht, Leland McInnes, John Healy, Charles-Antoine Dutertre,
  Immanuel~WH Kwok, Lai~Guan Ng, Florent Ginhoux, and Evan~W Newell.
\newblock Dimensionality reduction for visualizing single-cell data using
  {UMAP}.
\newblock \emph{Nature biotechnology}, 37\penalty0 (1):\penalty0 38--44, 2019.

\bibitem[Belkin \& Niyogi(2001)Belkin and Niyogi]{belkin2001laplacian}
Mikhail Belkin and Partha Niyogi.
\newblock Laplacian eigenmaps and spectral techniques for embedding and
  clustering.
\newblock \emph{Advances in neural information processing systems}, 14, 2001.

\bibitem[Carri{\'e}re et~al.(2021)Carri{\'e}re, Chazal, Glisse, Ike, Kannan,
  and Umeda]{carriere2021optimizing}
Mathieu Carri{\'e}re, Fr{\'e}d{\'e}ric Chazal, Marc Glisse, Yuichi Ike,
  Hariprasad Kannan, and Yuhei Umeda.
\newblock Optimizing persistent homology based functions.
\newblock In \emph{International Conference on Machine Learning}, pp.\
  1294--1303. PMLR, 2021.

\bibitem[Chari et~al.(2021)Chari, Banerjee, and Pachter]{chari2021specious}
Tara Chari, Joeyta Banerjee, and Lior Pachter.
\newblock The specious art of single-cell genomics.
\newblock \emph{bioRxiv}, 2021.

\bibitem[Chazal \& Michel(2017)Chazal and Michel]{chazal2017introduction}
Fr{\'e}d{\'e}ric Chazal and Bertrand Michel.
\newblock An introduction to topological data analysis: fundamental and
  practical aspects for data scientists.
\newblock \emph{arXiv preprint arXiv:1710.04019}, 2017.

\bibitem[Cherniavskii et~al.(2022)Cherniavskii, Tulchinskii, Mikhailov,
  Proskurina, Kushnareva, Artemova, Barannikov, Piontkovskaya, Piontkovski, and
  Burnaev]{cherniavskii2022acceptability}
Daniil Cherniavskii, Eduard Tulchinskii, Vladislav Mikhailov, Irina Proskurina,
  Laida Kushnareva, Ekaterina Artemova, Serguei Barannikov, Irina
  Piontkovskaya, Dmitri Piontkovski, and Evgeny Burnaev.
\newblock Acceptability judgements via examining the topology of attention
  maps.
\newblock \emph{Findings of the Association for Computational Linguistics:
  EMNLP 2022}, 2022.

\bibitem[Coenen \& Pearce(2019{\natexlab{a}})Coenen and
  Pearce]{coenen2019understanding}
Andy Coenen and Adam Pearce.
\newblock Understanding {UMAP}.
\newblock \emph{URL: https://pair-code. github. io/understanding-umap},
  2019{\natexlab{a}}.

\bibitem[Coenen \& Pearce(2019{\natexlab{b}})Coenen and Pearce]{mammoth1}
Andy Coenen and Adam Pearce.
\newblock Understanding {UMAP}, mammoth dataset.
\newblock 2019{\natexlab{b}}.
\newblock URL
  \url{https://github.com/PAIR-code/understanding-umap/tree/master/raw_data}.

\bibitem[Damrich \& Hamprecht(2021)Damrich and Hamprecht]{damrich2021umap}
Sebastian Damrich and Fred~A Hamprecht.
\newblock On umap's true loss function.
\newblock \emph{Advances in Neural Information Processing Systems},
  34:\penalty0 5798--5809, 2021.

\bibitem[Goodfellow et~al.(2016)Goodfellow, Bengio, Courville, and
  Bengio]{goodfellow2016deep}
Ian Goodfellow, Yoshua Bengio, Aaron Courville, and Yoshua Bengio.
\newblock \emph{Deep learning}, volume~1.
\newblock MIT press Cambridge, 2016.

\bibitem[Hinton \& Salakhutdinov(2006)Hinton and
  Salakhutdinov]{hinton2006reducing}
Geoffrey~E Hinton and Ruslan~R Salakhutdinov.
\newblock Reducing the dimensionality of data with neural networks.
\newblock \emph{science}, 313\penalty0 (5786):\penalty0 504--507, 2006.

\bibitem[Hofer et~al.(2019)Hofer, Kwitt, Niethammer, and
  Dixit]{hofer2019connectivity}
Christoph Hofer, Roland Kwitt, Marc Niethammer, and Mandar Dixit.
\newblock Connectivity-optimized representation learning via persistent
  homology.
\newblock In \emph{International conference on machine learning}, pp.\
  2751--2760. PMLR, 2019.

\bibitem[Karlov et~al.(2019)Karlov, Sosnin, Tetko, and
  Fedorov]{karlov2019chemical}
Dmitry~S Karlov, Sergey Sosnin, Igor~V Tetko, and Maxim~V Fedorov.
\newblock Chemical space exploration guided by deep neural networks.
\newblock \emph{RSC advances}, 9\penalty0 (9):\penalty0 5151--5157, 2019.

\bibitem[Kim et~al.(2020)Kim, Kim, Zaheer, Kim, Chazal, and
  Wasserman]{kim2020pllay}
Kwangho Kim, Jisu Kim, Manzil Zaheer, Joon Kim, Fr{\'e}d{\'e}ric Chazal, and
  Larry Wasserman.
\newblock Pllay: efficient topological layer based on persistent landscapes.
\newblock \emph{Advances in Neural Information Processing Systems},
  33:\penalty0 15965--15977, 2020.

\bibitem[Kingma \& Welling(2013)Kingma and Welling]{kingma2013auto}
Diederik~P Kingma and Max Welling.
\newblock Auto-encoding variational bayes.
\newblock \emph{arXiv preprint arXiv:1312.6114}, 2013.

\bibitem[Kobak \& Berens(2019)Kobak and Berens]{kobak2019art}
Dmitry Kobak and Philipp Berens.
\newblock The art of using {t-SNE} for single-cell transcriptomics.
\newblock \emph{Nature communications}, 10\penalty0 (1):\penalty0 1--14, 2019.

\bibitem[Kobak \& Linderman(2021)Kobak and Linderman]{kobak2021initialization}
Dmitry Kobak and George~C Linderman.
\newblock Initialization is critical for preserving global data structure in
  both {t-SNE} and {UMAP}.
\newblock \emph{Nature biotechnology}, 39\penalty0 (2):\penalty0 156--157,
  2021.

\bibitem[LeCun et~al.(1998)LeCun, Bottou, Bengio, and
  Haffner]{lecun1998gradient}
Yann LeCun, L{\'e}on Bottou, Yoshua Bengio, and Patrick Haffner.
\newblock Gradient-based learning applied to document recognition.
\newblock \emph{Proceedings of the IEEE}, 86\penalty0 (11):\penalty0
  2278--2324, 1998.

\bibitem[Leygonie et~al.(2021)Leygonie, Oudot, and
  Tillmann]{leygonie2021framework}
Jacob Leygonie, Steve Oudot, and Ulrike Tillmann.
\newblock A framework for differential calculus on persistence barcodes.
\newblock \emph{Foundations of Computational Mathematics}, pp.\  1--63, 2021.

\bibitem[Luo et~al.(2021)Luo, Xu, Zhang, and Jin]{luo2021topology}
Zixiang Luo, Chenyu Xu, Zhen Zhang, and Wenfei Jin.
\newblock A topology-preserving dimensionality reduction method for single-cell
  rna-seq data using graph autoencoder.
\newblock \emph{Scientific reports}, 11\penalty0 (1):\penalty0 1--8, 2021.

\bibitem[McInnes et~al.(2018)McInnes, Healy, and Melville]{mcinnes2018umap}
Leland McInnes, John Healy, and James Melville.
\newblock {UMAP}: Uniform manifold approximation and projection for dimension
  reduction.
\newblock \emph{arXiv preprint arXiv:1802.03426}, 2018.

\bibitem[Moon et~al.(2019)Moon, van Dijk, Wang, Gigante, Burkhardt, Chen, Yim,
  Elzen, Hirn, Coifman, et~al.]{moon2019visualizing}
Kevin~R Moon, David van Dijk, Zheng Wang, Scott Gigante, Daniel~B Burkhardt,
  William~S Chen, Kristina Yim, Antonia van~den Elzen, Matthew~J Hirn, Ronald~R
  Coifman, et~al.
\newblock Visualizing structure and transitions in high-dimensional biological
  data.
\newblock \emph{Nature biotechnology}, 37\penalty0 (12):\penalty0 1482--1492,
  2019.

\bibitem[Moor et~al.(2020)Moor, Horn, Rieck, and
  Borgwardt]{moor2020topological}
Michael Moor, Max Horn, Bastian Rieck, and Karsten Borgwardt.
\newblock Topological autoencoders.
\newblock In \emph{International conference on machine learning}, pp.\
  7045--7054. PMLR, 2020.

\bibitem[Nene et~al.(1996)Nene, Nayar, Murase, et~al.]{nene1996columbia}
Sameer~A Nene, Shree~K Nayar, Hiroshi Murase, et~al.
\newblock Columbia object image library (coil-100).
\newblock 1996.

\bibitem[Poulenard et~al.(2018)Poulenard, Skraba, and
  Ovsjanikov]{poulenard2018topological}
Adrien Poulenard, Primoz Skraba, and Maks Ovsjanikov.
\newblock Topological function optimization for continuous shape matching.
\newblock \emph{Computer Graphics Forum}, 37\penalty0 (5):\penalty0 13--25,
  2018.

\bibitem[Sainburg et~al.(2021)Sainburg, McInnes, and
  Gentner]{sainburg2021parametric}
Tim Sainburg, Leland McInnes, and Timothy~Q Gentner.
\newblock Parametric {UMAP} embeddings for representation and semisupervised
  learning.
\newblock \emph{Neural Computation}, 33\penalty0 (11):\penalty0 2881--2907,
  2021.

\bibitem[Szubert \& Drozdov(2019)Szubert and Drozdov]{szubert2019ivis}
Benjamin Szubert and Ignat Drozdov.
\newblock ivis: dimensionality reduction in very large datasets using siamese
  networks.
\newblock \emph{J. Open Source Softw.}, 4\penalty0 (40):\penalty0 1596, 2019.

\bibitem[Szubert et~al.(2019)Szubert, Cole, Monaco, and
  Drozdov]{szubert2019structure}
Benjamin Szubert, Jennifer~E Cole, Claudia Monaco, and Ignat Drozdov.
\newblock Structure-preserving visualisation of high dimensional single-cell
  datasets.
\newblock \emph{Scientific reports}, 9\penalty0 (1):\penalty0 1--10, 2019.

\bibitem[Tenenbaum et~al.(2000)Tenenbaum, De~Silva, and
  Langford]{tenenbaum2000global}
Joshua~B Tenenbaum, Vin De~Silva, and John~C Langford.
\newblock A global geometric framework for nonlinear dimensionality reduction.
\newblock \emph{Science}, 290\penalty0 (5500):\penalty0 2319--2323, 2000.

\bibitem[Tirosh et~al.(2016)Tirosh, Izar, Prakadan, Wadsworth, Treacy,
  Trombetta, Rotem, Rodman, Lian, Murphy, et~al.]{tirosh2016dissecting}
Itay Tirosh, Benjamin Izar, Sanjay~M Prakadan, Marc~H Wadsworth, Daniel Treacy,
  John~J Trombetta, Asaf Rotem, Christopher Rodman, Christine Lian, George
  Murphy, et~al.
\newblock Dissecting the multicellular ecosystem of metastatic melanoma by
  single-cell rna-seq.
\newblock \emph{Science}, 352\penalty0 (6282):\penalty0 189--196, 2016.

\bibitem[Van~der Maaten \& Hinton(2008)Van~der Maaten and
  Hinton]{van2008visualizing}
Laurens Van~der Maaten and Geoffrey Hinton.
\newblock Visualizing data using {t-SNE}.
\newblock \emph{Journal of machine learning research}, 9\penalty0 (11), 2008.

\bibitem[Wagner et~al.(2021)Wagner, Solomon, and Bendich]{wagner2021improving}
Alexander Wagner, Elchanan Solomon, and Paul Bendich.
\newblock Improving metric dimensionality reduction with distributed topology.
\newblock \emph{arXiv preprint arXiv:2106.07613}, 2021.

\bibitem[Wang et~al.(2021)Wang, Huang, Rudin, and
  Shaposhnik]{wang2021understanding}
Yingfan Wang, Haiyang Huang, Cynthia Rudin, and Yaron Shaposhnik.
\newblock Understanding how dimension reduction tools work: an empirical
  approach to deciphering t-{SNE}, {UMAP}, {TriMAP}, and {PaCMAP} for data
  visualization.
\newblock \emph{J Mach. Learn. Res}, 22:\penalty0 1--73, 2021.

\bibitem[Xiao et~al.(2017)Xiao, Rasul, and Vollgraf]{xiao2017online}
Han Xiao, Kashif Rasul, and Roland Vollgraf.
\newblock Fashion-mnist: a novel image dataset for benchmarking machine
  learning algorithms, 2017.

\bibitem[Yuan et~al.(2017)Yuan, Cai, Elowitz, Enver, Fan, Guo, Irizarry,
  Kharchenko, Kim, Orkin, et~al.]{yuan2017challenges}
Guo-Cheng Yuan, Long Cai, Michael Elowitz, Tariq Enver, Guoping Fan, Guoji Guo,
  Rafael Irizarry, Peter Kharchenko, Junhyong Kim, Stuart Orkin, et~al.
\newblock Challenges and emerging directions in single-cell analysis.
\newblock \emph{Genome biology}, 18\penalty0 (1):\penalty0 1--8, 2017.

\bibitem[Zhang et~al.(2020)Zhang, Xiao, and Wang]{zhang2020gpu}
Simon Zhang, Mengbai Xiao, and Hao Wang.
\newblock {GPU}-accelerated computation of vietoris-rips persistence barcodes.
\newblock In \emph{36th International Symposium on Computational Geometry}.
  Schloss Dagstuhl-Leibniz-Zentrum f{\"u}r Informatik, 2020.

\bibitem[Zomorodian(2001)]{zomorodian2001computing}
Afra~J. Zomorodian.
\newblock \emph{Computing and comprehending topology: Persistence and
  hierarchical {M}orse complexes ({P}h.{D}.{T}hesis)}.
\newblock University of Illinois at Urbana-Champaign, 2001.

\end{thebibliography}
\bibliographystyle{iclr2023_conference}

%----------------------------------------------------------
%
%      APPENDIX
%
%---------------------------------------------------------
\clearpage
\appendix
\section{Simplicial complexes and filtrations}
\label{app:tda_formal}

Here we briefly recall basic topological objects mentioned in our paper. Suppose we have a full graph $\mathrm{X} = \lbrace x_0, x_1, \ldots x_n\rbrace$ a set of points in some metric space $(R, d)$.

\textbf{Definition A.1} Any set $\sigma \subseteq X $ is \textit{(combinatorial) simplex}. Its \textit{vertices} are all points that belong to $K$. Its \textit{dimensionality} is number equal to $|\sigma| - 1$. Its \textit{faces} are all proper subsets of $\sigma$.

\textbf{Definition A.2} \textit{Simplicial complex} $\mathcal{C}$ is a set of simplices such that for every simplex $\sigma \in \mathcal{C}$ it contains all faces of $\sigma$ and for every two simplices $\sigma_1, \sigma_2 \in \mathcal{C}$ their intersection is face of both of them. 

Simplicial complexes can be seen as higher-dimensional generalization of graphs. There are many ways to build a simplical complex from a set of points, but only two important for this work: Vietoris-Rips and Chech complexes.

\textbf{Definition A.3} Given a threshold $\alpha$ the \textit{Vietoris-Rips (simplicial) complex} at threshold $\alpha$ (denotes as $\text{VR}_\alpha(\mathrm{X})$) is defined as set set of all simplices $\sigma$ such that $\forall x_i, x_j \in \sigma$ holds $d(x_i, x_j) \leq \alpha$. 

\textbf{Definition A.4} Given a threshold $\alpha$ the \textit{Čech (simplicial) complex} at threshold $\alpha$ (denotes as $\text{Cech}_\alpha(\mathrm{X})$) is defined as set set of all simplices $\sigma$ such that all closed balls of radius $\alpha$ and with centers in vertices of $\sigma$ have a non-empty intersection.

Alhough Čech complexes are rarely used in applications of Topological Data Analysis, they are important due to the fact that their fundamental topological properties are equal to those of the manifold `behind' $\mathrm{X}$ (so-called \textit{Nerve theorem}, see \citep{chazal2017introduction} for proper explanation).

The Vietoris-Rips complexes `approximate' Čech complexes :
$$
\text{VR}_\alpha(\mathrm{X}) \subseteq \text{Cech}_\alpha(\mathrm{X}) \subseteq \text{VR}_{2\alpha}(\mathrm{X})
$$

Note that the definition of the Vietoris-Rips complex doesn't require (even indirectly) function $d(.)$ to be metric - it should only be symmetric and non-negative. And so we can define the Vietoris-Rips complex of a weighted graph $G = (V, E)$. To do so we modify Definition A.3 by replacing $\mathrm{X}$ with $V$ and taking $d(v_i, v_j)$ as the weight of the edge between $v_i$ and $v_j$ for $i \neq j$ and $d(v_i, v_i) = 0, \forall i$.

In the scope of this work we consider only Vietoris-Rips complexes of graphs.

\textbf{Definition A.5} A \textit{filtration} of a simplicial complex  $\mathcal{C}$ is a nested family of subcomplexes $(\mathcal{C}_t)_{t \in T}$ , where 
$T \subseteq \mathrm{R}$,
such that for any $t_1, t_2 \in T$ , if $t_1 \leq t_2$ then $\mathcal{C}_{t_1} \subseteq \mathcal{C}_{t_2}$, and 
$\mathcal{C} = \bigcup_{t\in T} \mathcal{C}_{t}$. The set T may be
either finite or infinite.

Vietoris-Rips filtration can `reflect' topology of data set at every scale. Usually data sets are finite so there is finite number of thresholds that give different Vietoris-Rips complexes and thus finite filtration is enough for it.

\section{Formal definition of RTD}
\label{app:rtd_formal}

\begin{wrapfigure}{r}{0.3\textwidth}
    \centering
    \vskip-0.3in
    \includegraphics[width=0.3\textwidth]{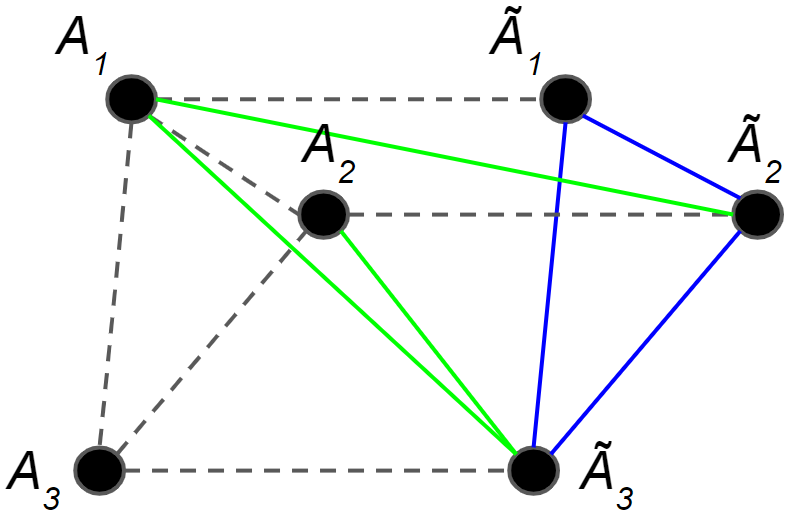}
    \caption{The graph $\hat{\mathcal{G}}^{w, \tilde{w}}$ to compare $\mathcal{G}^{w} = \{A_1, A_2, A_3\}$ and $\mathcal{G}^{\tilde{w}} = \{\tilde{A}_1, \tilde{A}_2, \tilde{A}_3\}$. Dashed edges correspond to zero weights, green edges to $w$, blue edges to $min(w, \tilde{w})$; edges with weight $+\infty$ are not shown.}
    \label{fig:double_graph}
     \vskip-0.4in
\end{wrapfigure}

The classic persistent homology is dedicated to the analysis of a single point cloud $X$. 
Recently, Representation Topology Divergence (RTD) \citep{barannikov2021representation} was proposed to measure the dissimilarity in the multi-scale topology between two point clouds $X, \tilde{X}$ of equal size $N$ with a one-to-one correspondence between clouds.

Let $\text{VR}_{\alpha}(\mathcal{G}^{w})$, $\text{VR}_{\alpha}(\mathcal{G}^{\tilde{w}})$ be two Vietoris-Rips simplicial complexes, where $w, \tilde{w}$ - are the distance matrices of $X, \tilde{X}$.
The idea behind RTD is to compare $\text{VR}_{\alpha}(\mathcal{G}^{w})$ with $\text{VR}_{\alpha}(\mathcal{G}^{min(w, \tilde{w}}))$, where $\mathcal{G}^{min(w, \tilde{w})}$ is the graph having weights $min(w, \tilde{w})$ on its edges. By definition, $\text{VR}_{\alpha}(\mathcal{G}^{w}) \subseteq \text{VR}_{\alpha}(\mathcal{G}^{min(w, \tilde{w})})$, $\text{VR}_{\alpha}(\mathcal{G}^{\tilde{w}}) \subseteq \text{VR}_{\alpha}(\mathcal{G}^{min(w, \tilde{w})})$.

To compare $\text{VR}_{\alpha}(\mathcal{G}^{w})$ with $\text{VR}_{\alpha}(\mathcal{G}^{min(w, \tilde{w}}))$, the auxiliary graph is constructed with doubled set of vertices $\hat{\mathcal{G}}^{w, \tilde{w}}$ (Figure \ref{fig:double_graph}) and weights on edges given in the simplest case by:
\begin{equation*}
%\label{min-matrix}
m= \begin{pmatrix}
  0 & (w_{+})^\intercal \\\
 w_{+} & \min(w, \tilde{w}) 
 \end{pmatrix},
\end{equation*}
where $w_{+}$ is the $w$ matrix with lower-triangular part replaced by $+\infty$, see (\citep{barannikov2021representation}, section 2.2) for the general form of the matrix $m$.
%Given two point clouds $X_1, X_2$ of equal size $N$, RTD introduces the weighted graph with $2N$ vertices.  Weights on edges are defined as
%$$
%m= \begin{pmatrix}
% 0 & (w_{1+})^\intercal \\\
% w_{1+} & \min(w_1,w_2) 
% \end{pmatrix},
%$$
%where $w_1, w_2$ are the matrices of pairwise distances of $X_1, X_2$ respectfully.
The persistence barcode of the weighted graph $\text{VR}(\hat{\mathcal{G}}^{w, \tilde{w}})$ is called the \textit{R-Cross-Barcode} (for \emph{Representations' Cross-Barcode}). Note that for every two nodes in the graph $\hat{\mathcal{G}}^{w, \tilde{w}}$ there exists a path with edges having zero weights. Thus, the $H_0$ barcode in the \textit{R-Cross-Barcode} is always empty.

Intuitively, the $k-$th barcode of $\text{VR}_\alpha(\hat{\mathcal{G}}^{w,\tilde{w}})$ records the 
 $k$-dimensional topological features that are born in $\text{VR}_{\alpha}(\mathcal{G}^{\min(w,\tilde{w}}))$ but are not yet born near the same place in $\text{VR}_{\alpha}(\mathcal{G}^{w})$,  and the $(k-1)-$dimensional topological features that are dead in $\text{VR}_{\alpha}(\mathcal{G}^{\min(w,\tilde{w})})$ but are not yet dead in $\text{VR}_{\alpha}(\mathcal{G}^{w})$.
The \emph{R-Cross-Barcode$_k (X,\tilde{X})$}
records the differences in the multi-scale topology of the two point clouds. The topological features with longer lifespans indicate in general the essential features.

Basic properties of {R-Cross-Barcode$_k(X,\tilde{X})$} (\cite{barannikov2021representation}) are: 
\begin{itemize}
\item if $X=\tilde{X}$, then for all $k$ {R-Cross-Barcode}$_k(X,\tilde{X})=\varnothing$;
\item if all distances within $\tilde{X}$ are zero i.e. all objects are represented  by the same point in $\tilde{X}$, then for all $k\ge0$: {R-Cross-Barcode$_{k+1}(X,\tilde{X})=\text{Barcode}_{k}(X)$} the standard barcode of the point cloud $X$;
\item for any value of threshold $\alpha$, the following sequence of natural linear maps of homology groups
\begin{multline}
    \xrightarrow{r_{3i+3}} H_{i}(VR_\alpha(\mathcal{G}^w)) \xrightarrow{r_{3i+2}} H_i(VR_\alpha(\mathcal{G}^{\min(w,\tilde{w})}))\xrightarrow{r_{3i+1}}\\ \xrightarrow{r_{3i+1}} H_i(VR_\alpha(\hat{\mathcal{G}}^{w,\tilde{w}})) \xrightarrow{r_{3i}} H_{i-1}(VR_\alpha(\mathcal{G}^w))\xrightarrow{r_{3i-1}}\\\xrightarrow{r_{3i-1}}\ldots\xrightarrow{r_1} H_{0}(VR_\alpha(\mathcal{G}^{\min(w,\tilde{w})}))\xrightarrow{r_0}0\label{eq:longseq}
\end{multline} is exact, i.e. for any $j$ the kernel of the map $r_{j}$ is the image of the map $r_{j+1}$.
\end{itemize}
\begin{proposition}
Given an exact sequence as in (\ref{eq:longseq}) with finite-dimensional filtered complexes $A_\alpha$, $B_\alpha$, $C_\alpha$, the alternating sums over  $k$ of their topological features lifespans satisfy
\begin{equation}
        \sum_k(-1)^{k} l_k(A)-\sum_k(-1)^{k} l_k(B)+\sum_k(-1)^{k} l_k(C)=0\label{eq:altsum}
        \end{equation} where $l_k(Z)$ denotes the sum of bars lengths in  {Barcode}$_{k}(Z)$, here for simplicity all lifespans are assumed to be finite. 
\end{proposition}
\begin{proof}
The exact sequence implies that the alternating sums of dimensions of the homology groups satisfy, for any $\alpha$, \begin{equation*}
        \sum_k(-1)^{k} \dim H_k(A_\alpha)-\sum_k(-1)^{k} \dim H_{k}(B_\alpha)+\sum_k(-1)^{k} \dim H_{k}(C_\alpha)=0
\end{equation*} 
Notice that for any $\alpha_1<\alpha_2$ 
\begin{equation*}
\dim H_k(Z_{\alpha_2}) -\dim H_k(Z_{\alpha_1})=\#b(Z,(\alpha_1,\alpha_2],k)- \#d(Z,(\alpha_1,\alpha_2],k) \end{equation*} where $\#b(Z,(\alpha_1,\alpha_2],k)$, respectfully $\#d(Z,(\alpha_1,\alpha_2],k)$, is the number of births, respectfully deaths, of dimension $k$ topological features in $Z$ at thresholds $\alpha$, $\alpha_1<\alpha\leq \alpha_2$. Hence 
\begin{multline}
        \sum_k(-1)^{k}(\#b- \#d)(A,(\alpha_1,\alpha_2],k) -\sum_k(-1)^{k} (\#b- \#d)(B,(\alpha_1,\alpha_2],k)+\\+\sum_k(-1)^{k} (\#b- \#d)(C,(\alpha_1,\alpha_2],k)=0  
\end{multline} 
Setting $\alpha_1=\alpha-\epsilon$, $\alpha_2=\alpha+\epsilon$,  we get, for any $\alpha$
\begin{equation*}
        \sum_k(-1)^{k}(\#b- \#d)(A,\alpha,k) -\sum_k(-1)^{k} (\#b- \#d)(B,\alpha,k)+\sum_k(-1)^{k} (\#b- \#d)(C,\alpha,k)=0  
\end{equation*} where $\#b(Z,\alpha,k)$, respectfully $\#d(Z,\alpha,k)$, is the number of  births, respectfully deaths, of dimension $k$ topological features in $Z$ at the threshold $\alpha$. Summing this over all nontrivial filtration steps $\alpha$ gives the identity (\ref{eq:altsum}).
%On the other hand for any $\alpha$
\end{proof}

%\hl{The exact sequence} (\ref{eq:longseq}) \hl{implies the following equality }for the alternating sum over all $k$ of {R-Cross-Barcode}$_k(X,\tilde{X})$ topological features' lifespans. 
\begin{proposition} 
%Let $l_k(v)$ denotes the sum of bars lengths in {Barcode}$_{k}(\mathcal{G}^v)$. Then  
\begin{equation}
        \sum_k(-1)^{k} RTD_k(w,\tilde{w})-\sum_k(-1)^{k} l_k(VR(\mathcal{G}^w))+\sum_k(-1)^{k} l_k(VR(\mathcal{G}^{\min(w,\tilde{w})})))=0
        \end{equation}
\end{proposition}

\section{Learning representations in higher dimensions}

\begin{table}[t]
\caption{Quality of data manifold global structure preservation at projection into 16D space.}
\centering
\label{tbl:real16dim-pca}
\begin{tabular}{lllll}
    \toprule
    & & \multicolumn{3}{c}{Quality measure}           \\
    \cmidrule(r){3-5}
    Dataset & Method   & L. C. & W. D. $H_{0}$ &  T. A. \\
    \midrule
    F-MNIST  & PCA & \textbf{0.977} & 351.3 $\pm$ 1.7 & \textbf{0.951 $\pm$ 0.005} \\
             & RTD-AE & \underline{0.960} & \textbf{181.2 $\pm$ 0.8} & \underline{0.907 $\pm$ 0.004} \\
    \midrule
   MNIST & PCA & \textbf{0.911} & 397.4 $\pm$ 1.3 & \textbf{0.863 $\pm$ 0.010} \\
         & RTD-AE & \underline{0.879} & \textbf{329.6 $\pm$ 2.6} & \underline{0.833 $\pm$ 0.006} \\
   \midrule
   COIL-20  & PCA & \textbf{0.966} & 196.4 $\pm$ 0.0 & \textbf{0.933 $\pm$ 0.004} \\
            & RTD-AE & \underline{0.944} & \textbf{88.9 $\pm$ 0.0} & \underline{0.892 $\pm$ 0.007} \\
   \bottomrule
\end{tabular}
\end{table}

The following table shows results of the experiment with latent dimensions  16, 32, 64 and 128 for the F-MNIST dataset. RTD-AE 
%and RTD-AE-MM (uses RTD modification, see Appendix \ref{app:rtd_mm}) 
are  consistently better than the competitors.

\begin{table}[t]
\caption{Quality of data manifold global structure preservation at projection into high-dimensional space.}
\label{tbl:dimhigh}
\centering
\begin{tabular}{lllllll}
    \toprule
     & & \multicolumn{5}{c}{Quality measure}           \\
     \cmidrule(r){3-7}
     Dataset & Method   & L. C. & W. D. $H_{0}$ & W. D. $H_{1}$ &  T. A. & RTD \\
     \midrule
     F-MNIST-128D & UMAP & 0.605 & 594.2 $\pm$ 3.0 & 20.73 $\pm$ 0.56 & 0.739 $\pm$ 0.010 & 12.33 $\pm$ 0.22\\
         & PCA & \textbf{0.996} & 107.4 $\pm$ 2.0 & 11.46 $\pm$ 0.39 & \textbf{0.981 $\pm$ 0.002} & 1.93 $\pm$ 0.08\\
         & PaCMAP & 0.589 & 587.9 $\pm$ 5.4 & 20.90 $\pm$ 0.27 & 0.736 $\pm$ 0.012 & 12.94 $\pm$ 0.51\\
         & Ivis & 0.521 & 551.7 $\pm$ 5.6 & 23.99 $\pm$ 0.69 & 0.693 $\pm$ 0.011 & 10.46 $\pm$ 0.23 \\
         & PHATE & 0.604 & 577.7 $\pm$ 3.8 & 21.57 $\pm$ 0.52 & 0.753 $\pm$ 0.010 & 10.61 $\pm$ 0.24 \\
         & AE & 0.892 & 240.5 $\pm$ 3.2 & 22.88 $\pm$ 1.14 & 0.860 $\pm$ 0.006 & 4.34 $\pm$ 0.15\\
         & TopoAE & 0.954 & 66.2 $\pm$ 2.1 & 12.02 $\pm$ 0.65 & \underline{0.902 $\pm$ 0.005} & 2.16 $\pm$ 0.14 \\
         & RTD-AE & 0.943 & \textbf{16.9 $\pm$ 1.8} & \textbf{9.43 $\pm$ 0.73} & 0.884 $\pm$ 0.010 & \textbf{1.41 $\pm$ 0.09}\\
        %  & RTD-AE-MM & \underline{0.964} & 32.7 $\pm$ 1.7 & \textbf{9.34 $\pm$ 0.86} & \underline{0.907 $\pm$ 0.004} &   \\
     \midrule
     F-MNIST-64D & UMAP & 0.596 & 590.7 $\pm$ 4.3 & 20.27 $\pm$ 0.71 & 0.735 $\pm$ 0.021 & 12.38 $\pm$ 0.35\\
         & PCA & \textbf{0.992} & 179.1 $\pm$ 1.9 & 18.61 $\pm$ 0.44 & \textbf{0.970 $\pm$ 0.003} & 3.10 $\pm$ 0.09 \\
         & PaCMAP & 0.510 & 590.5 $\pm$ 3.4 & 21.37 $\pm$ 0.47 & 0.731 $\pm$ 0.014 & 13.10 $\pm$ 0.37 \\
         & Ivis & 0.521 & 537.6 $\pm$ 3.3 & 26.86 $\pm$ 0.51 & 0.691 $\pm$ 0.011 & 10.34 $\pm$ 0.31 \\
         & PHATE & 0.586 & 586.1 $\pm$ 3.2 & 20.78 $\pm$ 0.52 & 0.751 $\pm$ 0.012 & 10.67 $\pm$ 0.36\\
         & AE & 0.888 & 281.0 $\pm$ 2.2 & 24.78 $\pm$ 0.86 & 0.861 $\pm$ 0.007 & 4.85 $\pm$ 0.18 \\
         & TopoAE & 0.938 & 89.3 $\pm$ 1.8 & 15.27 $\pm$ 0.68 & 0.889 $\pm$ 0.005 & 2.56 $\pm$ 0.13 \\
         & RTD-AE & 0.954 & \textbf{57.0 $\pm$ 0.6} & \textbf{11.76 $\pm$ 0.28} & \underline{0.895 $\pm$ 0.008} & \textbf{1.48 $\pm$ 0.09} \\
        %  & RTD-AE-MM & \underline{0.975} & 88.9 $\pm$ 0.6 & \textbf{12.16 $\pm$ 0.66} & \textbf{0.924 $\pm$ 0.004} &  \\
     \midrule
     F-MNIST-32D & UMAP & 0.593 & 597.1 $\pm$ 5.3 & 20.39 $\pm$ 0.24 & 0.741 $\pm$ 0.013 & 12.11 $\pm$ 0.30 \\
         & PCA & \textbf{0.986} & 263.0 $\pm$ 2.3 & 24.76 $\pm$ 0.97 & \textbf{0.960 $\pm$ 0.006} & 4.47 $\pm$ 0.12 \\
         & PaCMAP & 0.585 & 589.1 $\pm$ 4.9 & 21.15 $\pm$ 0.55 & 0.738 $\pm$ 0.010 & 12.61 $\pm$ 0.36 \\
         & Ivis & 0.696 & 559.8 $\pm$ 4.0 & 23.80 $\pm$ 0.57 & 0.770 $\pm$ 0.014 & 10.14 $\pm$ 0.29 \\
         & PHATE & 0.599 & 576.7 $\pm$ 3.5 & 21.79 $\pm$ 0.69 & 0.753 $\pm$ 0.011 & 10.48 $\pm$ 0.24\\
         & AE & 0.904 & 302.2 $\pm$ 2.6 & 26.37 $\pm$ 0.74 & 0.870 $\pm$ 0.008 & 5.28 $\pm$ 0.17 \\
         & TopoAE & 0.942 & 120.9 $\pm$ 2.5 & 15.84 $\pm$ 0.57 & 0.892 $\pm$ 0.006 & 2.49 $\pm$ 0.10\\
         & RTD-AE & 0.963 & \textbf{108.7 $\pm$ 1.8} & \textbf{14.03 $\pm$ 0.90} & \underline{0.907 $\pm$ 0.006} & \textbf{1.85 $\pm$ 0.06} \\
        %  & RTD-AE-MM & \underline{0.978} & 124.6 $\pm$ 1.4 & 14.60 $\pm$ 0.80 & \underline{0.929 $\pm$ 0.003} &  \\
     \midrule
     F-MNIST-16D & UMAP & 0.588 & 592.2 $\pm$ 4.0 & \textbf{20.37 $\pm$ 0.37} & 0.739 $\pm$ 0.013 & 12.31 $\pm$ 0.44 \\
         & PCA & \textbf{0.977} & 351.3 $\pm$ 1.7 & 29.15 $\pm$ 1.08 & \textbf{0.951 $\pm$ 0.005} & 5.91 $\pm$ 0.19 \\
         & PaCMAP & 0.600 & 585.9 $\pm$ 3.2 & 21.94 $\pm$ 0.59 & 0.741 $\pm$ 0.013 & 12.72 $\pm$ 0.48 \\
         & Ivis & 0.582 & 552.6 $\pm$ 3.5 & 24.83 $\pm$ 0.53 & 0.718 $\pm$ 0.014 & 10.76 $\pm$ 0.30\\
         & PHATE & 0.603 & 576.4 $\pm$ 4.4 & 21.61 $\pm$ 0.52 & 0.756 $\pm$ 0.016 & 10.72 $\pm$ 0.15\\
         & AE & 0.879 & 320.5 $\pm$ 1.9 & 27.01 $\pm$ 0.89 & 0.850 $\pm$ 0.004 & 5.52 $\pm$ 0.17 \\
         & TopoAE & 0.905 & 190.7 $\pm$ 1.2 & 25.65 $\pm$ 1.06 & 0.867 $\pm$ 0.006 & 3.69 $\pm$ 0.24 \\
         & RTD-AE & \underline{0.960} & \textbf{181.2 $\pm$ 0.8} & \textbf{20.94 $\pm$ 0.80} & \underline{0.907 $\pm$ 0.004} & \textbf{3.01 $\pm$ 0.13}\\
        %  & RTD-AE-MM & 0.943 & 197.2 $\pm$ 1.1 & 22.08 $\pm$ 1.14 & \underline{0.901 $\pm$ 0.004} & \\
    \bottomrule
\end{tabular}
\end{table}
%\fi

\section{Real world datasets, 2D latent space}
\label{app:real_world_2d}

\begin{figure}[t]
    \centering
    \includegraphics[width=\textwidth]{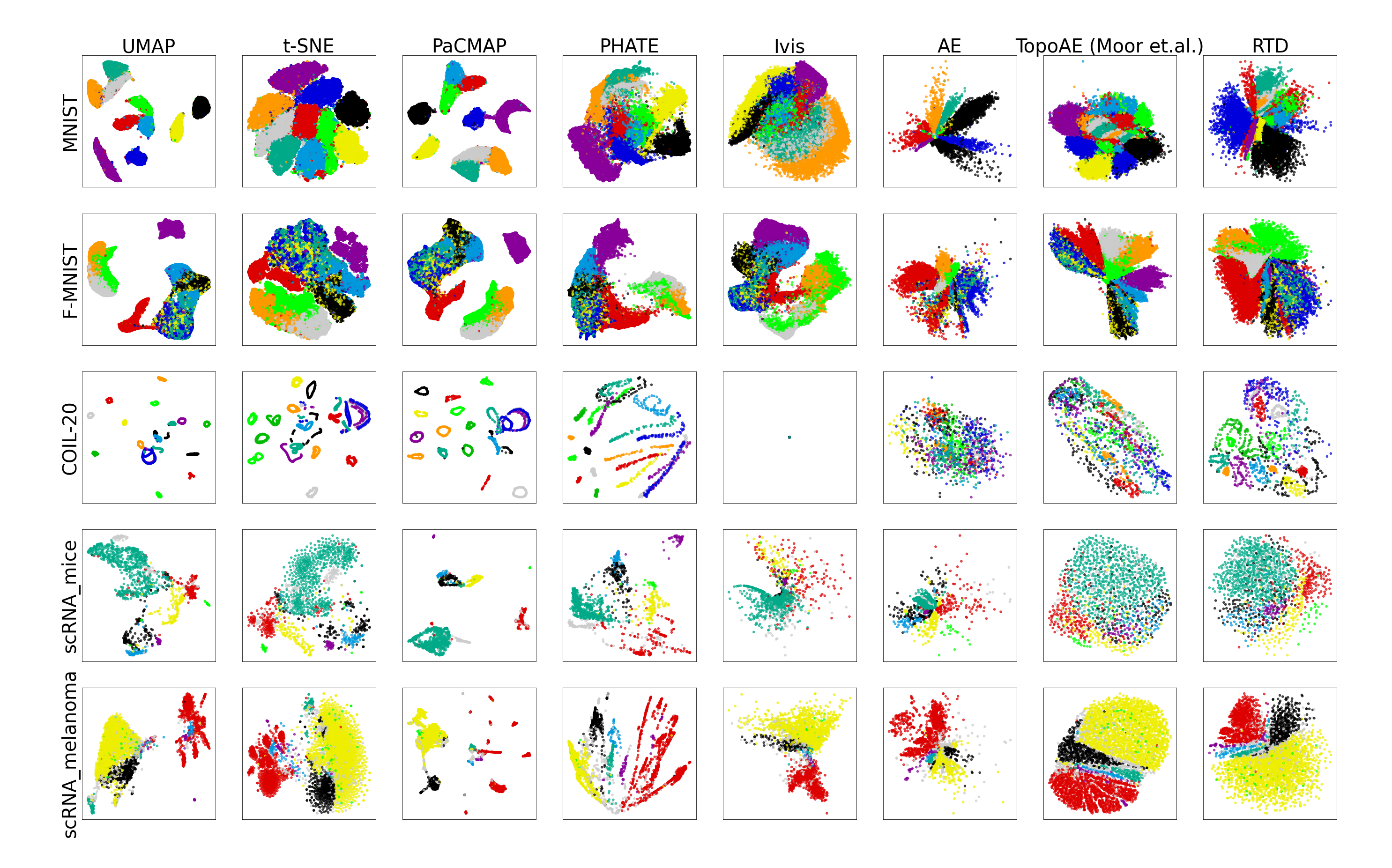}
    \caption{Results on real-world data reduction to 2D.}
    \label{fig:results_real}
\end{figure}

Table \ref{tbl:measure_real} and Figure \ref{fig:results_real} present the results.

\begin{table}
  \caption{Quality of data manifold global structure preservation for  real-world data dimension reduction to 2D.}
  \label{tbl:measure_real}
  \centering
  \begin{tabular}{lllllll}
    \toprule
    & & \multicolumn{4}{c}{Quality measure}           \\
    \cmidrule(r){3-6}
    Dataset & Method   & L. C. & W. D. $H_{0}$ & T. A. & RTD\\
    \midrule
    Mammoth & t-SNE     & 0.787          & 21.31 $\pm$ 0.25 & 0.830 $\pm$ 0.011    & 5.52 $\pm$ 0.12 \\
    & UMAP      & 0.776          & 28.64 $\pm$ 0.25 & 0.801 $\pm$ 0.016            & 6.81 $\pm$ 0.25\\
    & AE        & 0.966          & 21.94 $\pm$ 0.25 & 0.935 $\pm$ 0.005   & 6.38 $\pm$ 0.22 \\
    & PaCMAP    & 0.868         & 21.13 $\pm$ 0.21 &   0.866 $\pm$ 0.008           & 5.91 $\pm$ 0.29\\
    & Ivis      & 0.737          & \textbf{13.48 $\pm$ 0.30} &  0.764 $\pm$ 0.007  & 6.14 $\pm$ 0.20\\
    & TopoAE    & 0.915          & 21.51 $\pm$ 0.22 & 0.886 $\pm$ 0.007            & 5.16 $\pm$ 0.08\\
    & RTD-AE & \textbf{0.972} & 17.45 $\pm$ 0.23 & \textbf{0.928 $\pm$ 0.006} & \textbf{3.87 $\pm$ 0.07}\\
    \midrule
    F-MNIST & t-SNE & 0.547 & 602.9 $\pm$ 2.8 & 0.695 $\pm$ 0.011 & 11.11 $\pm$ 0.28 \\
    & UMAP & 0.595 & 616.5 $\pm$ 2.8 & 0.722 $\pm$ 0.011 & 11.72 $\pm$ 0.24 \\
    & AE & 0.762 & 614.7 $\pm$ 3.1 & 0.736 $\pm$ 0.012 & 11.51 $\pm$ 0.42\\
    & PaCMAP & 0.630 & 612.8 $\pm$ 6.0 & 0.732 $\pm$ 0.010 & 11.48 $\pm$ 0.27\\
    & Ivis & 0.496 & 609.2 $\pm$ 5.8 & 0.694 $\pm$ 0.011 & 11.70 $\pm$ 0.29\\
    & PHATE & 0.613 & 608.2 $\pm$ 2.7 & 0.739 $\pm$ 0.012 & 11.60 $\pm$ 0.22 \\
    & TopoAE & \textbf{0.795} & \textbf{599.0 $\pm$ 2.9} & \textbf{0.827 $\pm$ 0.011} & 11.84 $\pm$ 0.43\\
    & RTD-AE & 0.789 & \textbf{600.0 $\pm$ 3.1}  & 0.807 $\pm$ 0.011 & \textbf{10.67 $\pm$ 0.26}\\
    \midrule
    MNIST & t-SNE & 0.355 & 890.1 $\pm$ 6.8 & 0.611 $\pm$ 0.016 & \textbf{15.89 $\pm$ 0.28}\\
    & UMAP & 0.347 & 905.5 $\pm$ 6.7 & 0.612 $\pm$ 0.020 & 16.49 $\pm$ 0.21 \\
    & AE & 0.415 & 892.7 $\pm$ 5.9 & 0.635 $\pm$ 0.011 & 16.41 $\pm$ 0.27\\
    & PaCMAP & 0.310 & 902.8 $\pm$ 7.0 & 0.596 $\pm$ 0.015 & 16.41 $\pm$ 0.27 \\
    & Ivis & 0.377 & \textbf{887.5 $\pm$ 6.8} & 0.630 $\pm$ 0.014 & 16.03 $\pm$ 0.24\\
    & PHATE & 0.389 & 899.7 $\pm$ 3.4 & 0.623 $\pm$ 0.013 & 16.21 $\pm$ 0.29 \\
    & TopoAE & 0.349 & 891.5 $\pm$ 4.6 & 0.612 $\pm$ 0.014 & \textbf{15.71 $\pm$ 0.09} \\
    & RTD-AE & \textbf{0.501} & \textbf{885.1 $\pm$ 4.9} & \textbf{0.664 $\pm$ 0.009} & \textbf{15.79 $\pm$ 0.38}\\
    % & RTD-g  &  &  &  &  \\
    % & RTD-AE-MM & 0.818 & 468.578  & 14.821 & 0.83 & 0.661 \\
    % & RTD-L2  & 0.834 & 469.280 & 15.002 &  \textbf{0.84} & 0.652 \\
    % & RTD-AE-H2 & 0.714 & \textbf{249} & 30.1 \\
    \midrule
    COIL-20 & t-SNE & 0.462 & 273.9 $\pm$ 0.0  & 0.648 $\pm$ 0.025 & 12.23 $\pm$ 0.27 \\
        & UMAP & 0.247 & 279.5 $\pm$ 0.0 & 0.587 $\pm$ 0.013 & 13.72 $\pm$ 0.35 \\
        & AE & 0.667 & 271.8 $\pm$ 0.0 & 0.750 $\pm$ 0.016 & 11.82 $\pm$ 0.28 \\
        & PaCMAP & 0.506 & 276.6 $\pm$ 0.0 &  0.670 $\pm$ 0.012 & 12.60 $\pm$ 0.42 \\
        & Ivis & N.A. & N.A. & N.A. & N.A. \\
        & PHATE & 0.305 & 272.4 $\pm$ 0.000 & 0.592 $\pm$ 0.018 & 13.11 $\pm$ 0.39\\
        & TopoAE & 0.465 & \textbf{261.4 $\pm$ 0.0} & 0.662 $\pm$ 0.013 & 12.18 $\pm$ 0.23 \\
        & RTD-AE & \textbf{0.769} & \underline{262.9 $\pm$ 0.0} & \textbf{0.796 $\pm$ 0.009} & \textbf{11.51 $\pm$ 0.19}\\
    \midrule
    scRNA mice & t-SNE & 0.634 & 1151.3 $\pm$ 0.0 & 0.749 $\pm$ 0.010 & 21.93 $\pm$ 0.10\\
        & UMAP & 0.513 & 1161.3 $\pm$ 0.0 & 0.709 $\pm$ 0.015 & 22.26 $\pm$ 0.09\\
        & PCA & 0.733 & 1147.3 $\pm$ 0.0 & 0.790 $\pm$ 0.015 & 22.05 $\pm$ 0.06\\
        & AE & 0.677 & \textbf{1142.2 $\pm$ 0.0} & 0.778 $\pm$ 0.008 & 21.34 $\pm$ 0.17\\
        & PaCMAP & 0.483 & 1167.1 $\pm$ 0.0 & 0.693 $\pm$ 0.015 & 22.44 $\pm$ 0.08\\
        & Ivis & 0.254 & 1146.8 $\pm$ 0.0 & 0.602 $\pm$ 0.009 & 22.49 $\pm$ 0.25\\
        & PHATE & 0.522 & 1159.0 $\pm$ 0.0 & 0.711 $\pm$ 0.021 & 22.09 $\pm$ 0.07\\
        & TopoAE & 0.628 & 1144.2 $\pm$ 0.0 & 0.753 $\pm$ 0.019 & \textbf{21.15 $\pm$ 0.10} \\
        & RTD-AE & \textbf{0.780} & \textbf{1142.3 $\pm$ 0.0} & \textbf{0.797 $\pm$ 0.010} & \textbf{21.03 $\pm$ 0.13} \\
    \midrule
    scRNA melanoma & t-SNE & 0.505 & 1445.6 $\pm$ 3.2 & 0.699 $\pm$ 0.019 & 20.45 $\pm$ 0.38\\
        & UMAP & 0.471 & 1459.5 $\pm$ 3.0 & 0.684 $\pm$ 0.014 & 20.87 $\pm$ 0.43\\
        & PCA & 0.536 & 1446.7 $\pm$ 3.1 & 0.722 $\pm$ 0.014 & 20.64 $\pm$ 0.39\\
        & AE & 0.407 & 1442.0 $\pm$ 3.5 & 0.684 $\pm$ 0.014 & 20.90 $\pm$ 0.41\\
        & PaCMAP & 0.401 & 1460.9 $\pm$ 3.1 & 0.674 $\pm$ 0.017 & 20.91 $\pm$ 0.38 \\
        & Ivis & 0.504 & 1442.5 $\pm$ 3.1 & 0.699 $\pm$ 0.016 & 20.50 $\pm$ 0.35\\
        & PHATE & 0.427 & 1458.8 $\pm$ 3.1 & 0.689 $\pm$ 0.023 & 20.94 $\pm$ 0.41\\
        & TopoAE & 0.521 & 1442.9 $\pm$ 3.2 & 0.709 $\pm$ 0.009 & 20.23 $\pm$ 0.38 \\
        & RTD-AE & \textbf{0.639} & \textbf{1438.2 $\pm$ 3.0} & \textbf{0.747 $\pm$ 0.009} & \textbf{19.81 $\pm$ 0.37} \\
    \bottomrule
  \end{tabular}
\end{table}

\section{Synthetic datasets, 2D latent space}
\label{app:synthetic_2d}

Table \ref{tbl:measure_synthetic} shows the results.

\begin{table}[h!]
  \caption{Quality of data manifold global structure preservation on synthetic data.}
  \label{tbl:measure_synthetic}
  \centering
  \begin{tabular}{lllllll}
    \toprule
    & & \multicolumn{5}{c}{Quality measure}           \\
    \cmidrule(r){3-7}
    Dataset & Method   & L. C. & W. D. $H_{0}$ & W.D. $H_{1}$ & T. A. & RTD \\
    \midrule
    Circle & t-SNE & 0.986 & 1.073 & 0.079 & 0.95 & 0.59\\
            & UMAP & 0.808 & 1.823 & 0.712 & 0.81 & 1.46 \\
            & AE & 0.630 & 1.179 & 0.744 & 0.81 & 1.02 \\
            % & {\color{blue} PCA} & {\color{blue}\textbf{1.0}} & {\color{blue}0.0} & {\color{blue}1.0} \\
            % & {\color{blue} MDS} & {\color{blue}\textbf{1.0}} & {\color{blue}0.011} & {\color{blue}0.998} \\
            & PaCMAP & 0.747 & 2.263 & N.A. & 0.81 & 1.61 \\
            & Ivis & \textbf{0.990} & 0.182 & N.A. & \textbf{0.96} & 0.18 \\
            & PHATE & 0.891 & 0.871 & N.A. & 0.88 & 1.04 \\
            & TopoAE & 0.978 & 0.220 & 0.080 & 0.95 & 0.19\\
            & RTD-AE & 0.984 & \textbf{0.105} & \textbf{0.070} & \textbf{0.96} & \textbf{0.07} \\
            % & RTD-g  & \underline{0.983} & \textbf{0.069} & \textbf{0.063} & 0.95 & \\
            % & RTD-gMM & 0.973 & 0.101 & 0.105 & 0.94 & \\
            % & RTD-gL2  & 0.914 & 0.194 & 0.254 & 0.89 & \\
            % & RTD-AE-H2   & \textbf{0.973} & \textbf{0.372} & \textbf{0.109} \\
    \midrule
    2 Clusters & t-SNE & 0.633 & 9.122 & 1.171 & 0.72 & 5.18 \\
        & UMAP & 0.542 & 9.003 & 0.914 & 0.84 & 6.27 \\
        & AE & 0.925 & 1.654 & 0.807 & \textbf{0.94} & 2.03 \\
        % & {\color{blue} PCA} & {\color{blue}\textbf{1.0}} & {\color{blue}0.0} & {\color{blue}1.0} \\
        % & {\color{blue} MDS} & {\color{blue}\textbf{1.0}} & {\color{blue}0.106} & {\color{blue}0.999} \\
        & PaCMAP & 0.269 & 10.41 & N.A. & 0.64 & 5.71 \\
        & Ivis & 0.423 & 7.400 & N.A. & 0.76 & 5.58\\
        & PHATE & 0.281 & 7.356 & N.A. & 0.66 & 4.90 \\
        & TopoAE & 0.719 & 7.692 & 0.883 & 0.87 & 3.59 \\
        & RTD-AE & \textbf{0.999} & \textbf{0.313} & \textbf{0.313} & \textbf{0.96} & \textbf{0.32} \\
        % & RTD-g  & \textbf{0.999} & 0.533 & 0.352 & \textbf{0.99}  \\
        % & RTD-gMM & \textbf{0.999} & \textbf{0.451} & \textbf{0.335} & \textbf{0.99}  \\
        % & RTD-gL2  & 0.994 & 0.861 & 0.394 & 0.98  \\
        % & RTD-AE-H2 & \textbf{0.999} & \textbf{1.267} & \textbf{0.896} \\
    \midrule
    3 Clusters & t-SNE & 0.751 & 4.111 & 0.370 & 0.81 & 0.91\\
        & UMAP & 0.615 & 2.671 & 0.280 & 0.78 & 0.83 \\
        & AE & 0.907 & 1.013 & 0.054 & 0.93 & 0.59 \\
        % & {\color{blue} PCA} & {\color{blue}\textbf{1.0}} & {\color{blue}0.0} & {\color{blue}1.0} \\
        % & {\color{blue} MDS} & {\color{blue}\textbf{1.0}} & {\color{blue}0.014} & {\color{blue}0.996} \\
        & PaCMAP & 0.778 & 2.620 & N.A. & 0.89 & 0.92 \\
        & Ivis & 0.918 & 2.511 & N.A. & 0.82 & 1.30 \\
        & PHATE & 0.651 & 1.538 & N.A. & 0.72 & 0.99 \\
        & TopoAE & 0.997 & 0.586 & 0.054 & 0.81 & 0.13 \\
        & RTD-AE & \textbf{0.999} & \textbf{0.307} & \textbf{0.028} & \textbf{0.99} & \textbf{0.11} \\
        % & RTD-g & 0.997 & \textbf{0.567} & \textbf{0.038} & 0.94 \\
        % & RTD-gMM & \textbf{0.999} & \underline{0.570} & \textbf{0.038} & 0.94 \\
        % & RTD-gL2  & \underline{0.998} & 0.578 & \textbf{0.038} & 0.94 \\
        % & RTD-AE-H2 & \textbf{0.999} & 1.853 & \textbf{0.109} \\
    \midrule
    Random & t-SNE & 0.981 & 4.182 & 1.938 & 0.95 & 1.54\\
        & UMAP & 0.950 & 0.979 & 0.622 & 0.91 & 0.55 \\
        & AE & 0.700 & 9.976 & 2.343 & 0.75 & 1.32\\
        % & {\color{blue} PCA} & {\color{blue}\textbf{1.0}} & {\color{blue}0.0} & {\color{blue}1.0} \\
        % & {\color{blue} MDS} & {\color{blue}\textbf{1.0}} & {\color{blue}0.042} & {\color{blue}0.999} \\
        & PaCMAP & 0.982 & 5.398 & N.A. & 0.95 & 2.08 \\
        & Ivis & 0.648 & 11.49 & N.A. & 0.75 & 2.17 \\
        & PHATE & 0.945 & 6.703 & N.A. & 0.92 & 2.12\\
        & TopoAE & 0.854 & 3.288 & 1.367 & 0.84 & 0.91\\
        & RTD-AE & \textbf{0.996} & \textbf{0.148} & \textbf{0.389} & \textbf{0.98} & \textbf{0.17}\\
        % & RTD-g  & 0.997 & 0.143 & 0.264 & \textbf{0.99} \\
        % & RTD-gMM & \textbf{0.998} & \textbf{0.121} & \textbf{0.242} & \textbf{0.99} \\
        % & RTD-gL2  & \textbf{0.998} & 0.330 & 0.367 & 0.98 \\
        % & RTD-AE-H2 & \textbf{0.992} & 2.212 & 1.27 \\
    \bottomrule
  \end{tabular}
\end{table}

\section{RTD minimization without the autoencoder}
\label{app:pure_rtd}

\begin{figure*}[tbp]
\centering
\begin{subfigure}{0.2\textwidth}
\includegraphics[width=\textwidth]{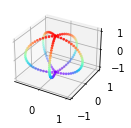}
\caption{Original 3D data}
\end{subfigure}
\begin{subfigure}{0.2\textwidth}
\includegraphics[width=\textwidth]{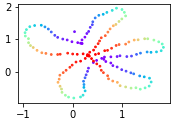}
\caption{RTD}
\end{subfigure}
\begin{subfigure}{0.2\textwidth}
\includegraphics[width=\textwidth]{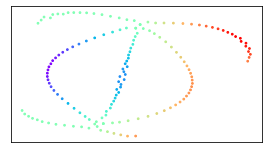}
\caption{t-SNE}
\end{subfigure}
\begin{subfigure}{0.2\textwidth}
\includegraphics[width=\textwidth]{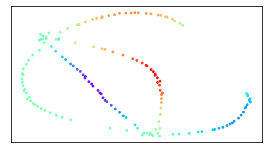}
\caption{UMAP}
\end{subfigure}
\caption{``Meridians on the sphere'' dataset. Notice the disconnectedness of meridians in (c) and (d).}
\label{fig:meridians}
\end{figure*}

\begin{figure*}[tbp]
\centering
\begin{subfigure}{0.2\textwidth}
\includegraphics[width=\textwidth]{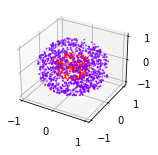}
\caption{Original 3D data}
\end{subfigure}
\begin{subfigure}{0.2\textwidth}
\includegraphics[width=\textwidth]{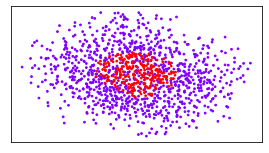}
\caption{RTD}
\end{subfigure}
\begin{subfigure}{0.2\textwidth}
\includegraphics[width=\textwidth]{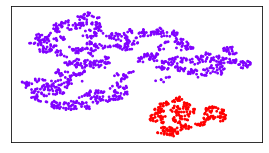}
\caption{t-SNE}
\end{subfigure}
\begin{subfigure}{0.2\textwidth}
\includegraphics[width=\textwidth]{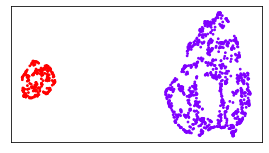}
\caption{UMAP}
\end{subfigure}
\caption{``Nested spheres'' dataset.}
\label{fig:spheres}
\end{figure*}

Given the set $X = \{x_i\}_{i=1}^n$ of $n$ objects in high-dimensional space $x_i \in \mathbb{R}^d$, our goal is to find their representations in low-dimensional space $Z = \{ z_i\}$, $z_i \in \mathbb{R}^k$.
It is possible to solve
$$
\min_{Z} \text{RTD}(X, Z)
$$
directly w.r.t $n$ vectors $z_i \in \mathbb{R}^k$, in the flavor similar to UMAP and t-SNE. 
Figures \ref{fig:meridians}, \ref{fig:spheres} show the results of two experiments with 3D$\to$2D dimensionality reduction. We conclude that dimensionality reduction via RTD optimization better preserves data topology: meridians are kept connected (Figure \ref{fig:meridians}) and the nestedness is retained (Figure \ref{fig:spheres}). The optimization took $\sim$1 hour. For the experiment with nested spheres, the RTD optimization was warmstarted with the MDS solution.

\section{Alternative RTD variant}
\label{app:rtd_mm}

RTD relies on the auxiliary graph with doubled set of vertices $\hat{\mathcal{G}}^{w, \tilde{w}}$ and weights on edges:
\[
\label{min-matrix}
m= \begin{pmatrix}
  0 & (w_{+})^\intercal \\\
 w_{+} & \min(w, \tilde{w}) 
 \end{pmatrix}.
\]

An alternative variant of RTD is possible with the following matrix of weights:
\[
m= \begin{pmatrix}
  0 & \text{max}(w, \tilde{w})_{+}^\intercal \\\
 \text{max}(w, \tilde{w})_{+} & w 
 \end{pmatrix}
\]
 in the simplest case.
Both of them share similar properties and guarantee that $\text{RTD}(X,Z)=0$ when all the pairwise distances in point clouds $X$ and $Z$ are the same. Also in both cases if $RTD_k(X,Z)=RTD_k(Z,X)=0$ for $k\geq 1$ then the persistence diagrams of $X$ and $Z$ coincide. The minimization of the sum of both variants of RTD leads to richer gradient information. We used this loss in the experiment with the ``Mammoth'' dataset. 
%and in the F-MNIST projected to dimensions 16-128 experiment where it was denoted RTD-AE-MM.

\section{Hyperparameters}
\label{app:hyperparams}

%In the experiment with the ``Mammoth'' dataset we used autoencoder with hidden layers of size 32,32,2,32,32 and ReLU activations. We used ADAM with learning rate $10^{-3}$ and batch size 256. We initially trained autoencoder for 5 epochs with the reconstruction loss, then continued with the RTD loss only for 50 epochs. 

In the experiments with projecting to 3D-space we trained model for 100 epochs using Adam optimizer.  We initially trained autoencoder for 10 epochs with only the reconstruction loss and learning rate 1e-4, then continued with RTD. Epochs 11-30 were trained with learning rate 1e-2, epochs 31-50 with learning rate 1e-3 and for epochs all after learning rate 1e-4 was used. Batch size was 80.

For 2D and high-dimensional projections, we used fully-connected autoencoders with hyperparameters specified in the Table \ref{tbl:hyperparams_description}. The autoencoder was initially trained only with reconstruction loss for some number of epochs, and then the RTD loss kicked in. The learning rate stayed the same for an entire duration of training.

For experiments we used NVIDIA TITAN RTX.

\begin{table}[ht!]
    \centering
    \caption{Hyperparameters description.}
    \label{tbl:hyperparams_description}
    \begin{tabular}{lllllll}
        \toprule
        Dataset name & Batch size & LR & Hidden dim & \# layers & Epochs & RTD epoch \\
        \toprule
        Circle & $80$ & $10^{-3}$ & $16$ & $3$ & $100$ & $20$\\
        \midrule
        Random & $80$ & $10^{-3}$ & $16$ & $3$ & $100$ & $20$ \\
        \midrule
        2 Clusters & $80$ & $10^{-3}$ & $16$ & $3$ & $100$ & $20$\\
        \midrule
        3 Clusters & $80$ & $10^{-3}$ & $16$ & $3$ & $100$ & $20$\\
        % \midrule
        % Random10D & $80$ & $10^{-3}$ & $32$ & $3$ & $100$ & $30$ \\
        \midrule
        Mammoth & $256$ & $10^{-3}$ & $32$ & 3 & $100$ & $5$\\
        \midrule
        MNIST & $256$ & $10^{-4}$ & $512$ & $3$ & $250$ & $60$\\
        \midrule
        F-MNIST & $256$ & $10^{-4}$ & $512$ & $3$ & $250$ & $60$\\
        \midrule
        COIL-20 & $256$ & $10^{-4}$ & $512$ & $3$ & $250$ & $60$\\
        \midrule
        scRNA mice & $256$ & $10^{-3}$ & $768$ & $3$ & $250$ & $60$\\
        \midrule
        scRNA melanoma & $256$ & $10^{-3}$ & $768$ & $3$ & $250$ & $60$\\
        % \midrule
        % Calorimeter & $256$ & $10^{-4}$ & $512$ & $6$ & $250$ & $60$\\
        \bottomrule
    \end{tabular}
\end{table}
% Eduard
\section{RTD optimization speed-ups}
\label{app:rtd_tricks}
%% _______________________________________________________________________________________
%% NOTE: this section is about optimization heuristics included in the code distributed on 5th of April. Previous version doesn't implement any of them.
%% This should be removed if we decide to use only old version of the code
%% _______________________________________________________________________________________

For all computations of RTD-barcodes in this work we used modified version of  Ripser++ software \citep{zhang2020gpu}. Modification that we made was intended at decreasing computational time via exploration of the structure of graph $\hat{\mathcal{G}}^{w, \tilde{w}}$ (see Section 3.2). The idea behind it is to reduce the size of filtered complex by excluding from it the simplices that do not affect the persistence homology.

Here we  consider only simplices of dimension at least 1.

We exclude all simplices spanned by vertices from the first half of the vertex set of the graph $\hat{\mathcal{G}}^{w, \tilde{w}}$.  Those are the vertices corresponding to the upper-left quadrant of the graph`s edge weights matrix $m$ from section \ref{app:rtd_formal}. All of them have diameters equal to zero.  And if any such simplex spawn a topological feature, it is immediately killed by another such simplex.

As before, let $N$ be the number of vertices in point clouds. Then $\hat{\mathcal{G}}^{w, \tilde{w}}$ has $2N$ vertices and our modification eliminates $\binom{N}{d}$ out of $\binom{2N}{d}$ simplices of dimension $d - 1$.

In particular, this eliminates around $1/8$ of rows and $1/4$ of columns (around $1/3$  cells in total) from the boundary matrix used for the computation of persistence pairs of dimension 1. On average, comparing to the standard Ripser++ computation, this gives  $\approx 45\%$ less time  for the computation of  persistence intervals of dimension 1.

Next, we  describe some techniques that can improve convergence when RTD is to be minimized without an autoencoder (\ref{app:pure_rtd}).

Usually we perform (sub)gradient descent to minimize $\text{RTD}(X, \tilde{X})$ between ``movable'' cloud $X$ and given constant $\tilde{X}$. 

\paragraph{Gradient smoothing.} Subgradients computed at each step of this procedure associate each homological class with at most $4$ points from $X$, while topological structures often include much more. Moreover, adjustments w.r.t. them may be inconsistent for nearby points.

To overcome this, we ``smooth'' gradients by passing to each point averaged gradients of all its neighbours. Let $\nabla_i^{(k)} $ be the gradient value for $X_i$ at step $k$ and $U(X_i^{(k)})$ be some neighbourhood of $X_i^{(k)}$. Then the formula for each step of the gradient descent is
$$
X_i^{(k)} = X_i^{(k - 1)} - \lambda_{k} \left(\beta\nabla_i^{(k)} + (1 - \beta)\frac{1}{\#\lbrace X_j^{(k)} \in U(X_i^{(k)}) \rbrace}\sum_{X_j^{(k)} \in U(X_i^{(k)})}   \nabla_j^{(k)}\right) 
$$
Here $\beta \in [0; 1]$ is some parameter.

\paragraph {Minimum bypassing.}  Suppose we want to shorten an edge $m_{i + N, j + N}$ from bottom-right quadrant of matrix $m$ (i.e. $\frac{\partial ~ \text{RTD}(X, \tilde{X})}{\partial m_{i + N, j + N}} < 0 $). 
It may occur that $w_{i, j} > \tilde{w}_{i, j}$, so 
$$\frac{\partial m_{i + N, j + N}}{\partial X_{i}} = \frac{X_{i} - X_{j}}{\lvert\rvert X_i - X_{j}\lvert\rvert_2}\mathbb{I}\lbrace w_{i, j} < \tilde{w}_{i, j}\rbrace = 0 $$
and gradient descent will stuck here (since $\tilde{X}$ is constant). Thus there may appear a certain threshold below which $\text{RTD}(X, \tilde{X})$ can't be minimized in this case. But it \textit{can} be further minimized if we move points $X_{i}$ and $X_{j}$ close enough to each other so $w_{i, j} < \tilde{w}_{i, j}$.

To do it, if $\frac{\partial ~ \text{RTD}(X, \tilde{X})}{\partial m_{i + N, j + N}} < 0$ , we compute $\frac{\partial m_{i, j}}{\partial X_{i}}$ without indicator, i.e. as $\frac{X_{i} - X_{j}}{\lvert\rvert X_i - X_{j}\lvert\rvert_2}$. This will assure $w_{i, j}$ is decreasing and at certain point will became lower than $\tilde{w}_{i, j}$.

If $\frac{\partial ~ \text{RTD}(X, \tilde{X})}{\partial m_{i + N, j + N}} \geq 0$ we don't change anything, because the discussing effect appears only if we minimize a minimum of a function and a constant.

We performed an experiment to transform a cloud in the shape of the infinity sign  by minimizing the  RTD between this cloud and a ring-shaped cloud. Both clouds had 100 points and we did not use batch optimisation. We performed 100 iterations of gradient descend to minimize RTD in each of the following four setups: using none, each or both of Gradient Smoothing and Minimum Bypassing tricks. For each setup we also searched for the best learning rate. The Table \ref{tbl:rtd-speedups} shows the results after 100 iterations.

\begin{table}[t]
  \caption{{RTD optimization speed-ups}}
  \label{tbl:rtd-speedups}
  \centering
  \begin{tabular}{lcc}
    \toprule
    Optimization trics   & RTD  & Relative value ($\%$)  \\
    \midrule
    None &  $3.09059$  & $100.00\%$  \\
    Gradient Smoothing &  $2.68406$ & $86.73\%$ \\
    Minimum Bypassing & $1.64366$ & $53.07\%$ \\
    Both & $1.20738$ & $38.84\%$  \\
    \bottomrule
  \end{tabular}
\end{table}

\section{Comparison with TopoAE loss}
\label{app:topoae_critique}

\begin{figure*}[th!]
    \centering
    \includegraphics[width=0.8\linewidth]{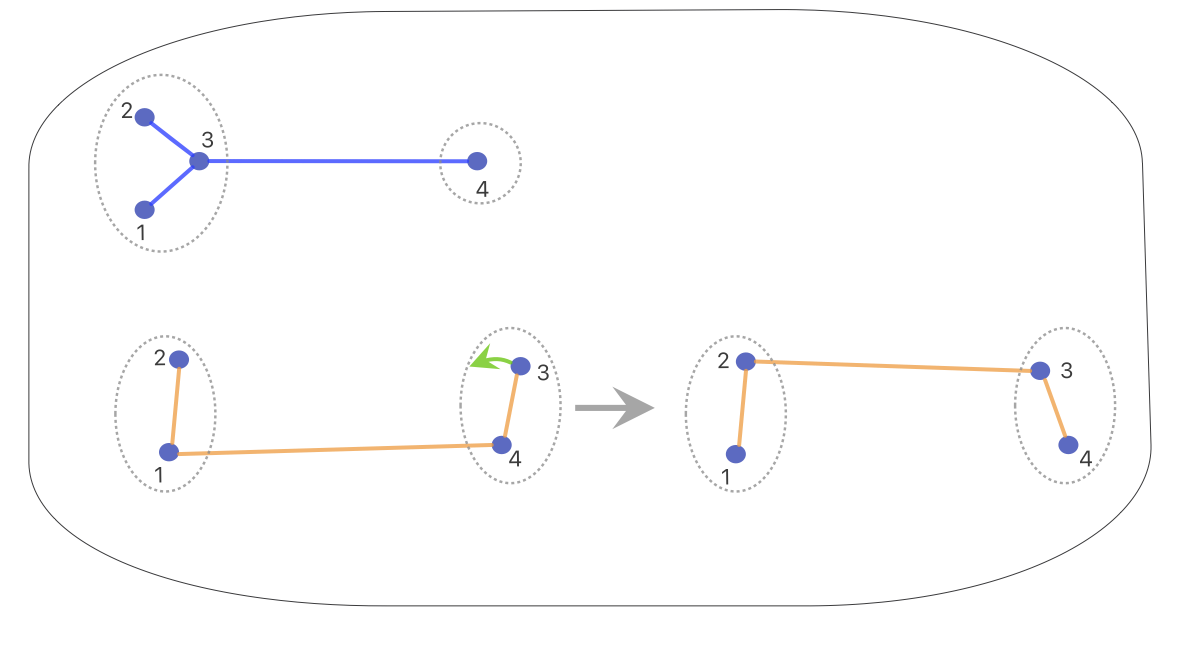}
    \caption{Discontinuity of the TopoAE loss. The point cloud $\tilde{X}$  consists of two clusters $\{1,2,3\}$ and $\{4\}$ (top). The point cloud ${X}$ (bottom left) consists of two clusters $\{1,2\}$ and $\{3,4\}$. (bottom left).  The distances within each cluster are  of order 
$10^{-1}$ and the distances between the clusters equal to $10^3\pm 10^{-1}$. The TopoAE loss is discontinuous because under a small perturbation of points, the minimal spanning tree $\Gamma$ may change. When the point $3$ moves slightly as indicated, then the minimal spanning tree $\Gamma$, coloured by yellow, changes and the term  $(w_{14}-\tilde{w}_{14})^2\sim 10^{-2}$  in TopoAE loss is replaced by $(w_{23}-\tilde{w}_{23})^2\sim 10^6$.  }
    \label{fig:TopAEbreak}
\end{figure*}

The following simple example on Figure \ref{fig:TopAEbreak} shows that the
TopoAE loss can be discontinuous in a rather standard situation. The TopoAE loss \citep{moor2020topological} is constructed by calculating first the two minimal spanning trees $\Gamma$, $\tilde{\Gamma}$ for each of the graphs $\mathcal{G}^w$, $\mathcal{G}^{\tilde{w}}$, whose weights are the distances within two point clouds $X$ and $\tilde{X}$. Then the TopoAE loss is the sum of two terms $L^{\text{TopoAE}}=l+\tilde{l}$. One term is the sum over the set of edges of $\Gamma$: $l=\frac{1}{2}\sum_{ij\in\text{Edges}(\Gamma)}(w_{ij}-\tilde{w}_{ij})^2$, and the other is the analogous sum over the edges of $\tilde{\Gamma}$: $\tilde{l}=\frac{1}{2}\sum_{ij\in\text{Edges}(\tilde{\Gamma})}(w_{ij}-\tilde{w}_{ij})^2$. Under a small perturbation of points, the minimal spanning tree $\Gamma$ may change, e.g. with a change of pair of the closest points from two clusters. But then the corresponding weights $\tilde{w}$ change in general discontinuosly.
The point cloud $\tilde{X}$ on Figure \ref{fig:TopAEbreak}  consists of two clusters $\{1,2,3\}$ and $\{4\}$. The point cloud ${X}$ consists of two clusters $\{1,2\}$ and $\{3,4\}$. We set the distances within each cluster to be  of order 
$10^{-1}$ and the distance between the clusters equal to $10^3\pm 10^{-1}$. When the point $3$ moves in $X$ slightly as indicated, then the minimal spanning tree $\Gamma$, coloured by yellow, changes and the term $(w_{14}-\tilde{w}_{14})^2\sim 10^{-2}$ in $l$ is replaced by $(w_{23}-\tilde{w}_{23})^2\sim 10^6$.

\begin{proposition}
The RTD loss is continuous. The $RT\!D_k(X, \tilde{X})$  depends continuously on $(X, \tilde{X})$. 
\end{proposition}
The proof follows from the stability of the barcode of the filtered complex $\text{VR}_\alpha(\hat{\mathcal{G}}^{w,\tilde{w}})$ with respect to the bottleneck distance under perturbation of the edge weights, see Appendix \ref{sec:stability}.
\section{Stability of  \textit{R-Cross-Barcode} and \textit{RTD}}\label{sec:stability}

\begin{proposition}
{For any perturbations $X'$ of a point cloud $X$ and $\tilde{X}'$ of a point cloud $\tilde{X}$, }
\begin{multline}
    d_B(\text{R-Cross-Barcode}_k(X,\tilde{X}),\text{R-Cross-Barcode}_k(X', \tilde{X}'))\\ \leq 2\max(\max_i \lVert X'_i-X_i\rVert ,\max_j\lVert \tilde{X}'_j-\tilde{X}_j\rVert) 
\end{multline} {where $d_B$ denotes the bottleneck distance.}
\end{proposition}
\begin{proof}
    {By construction, the $\text{R-Cross-Barcode}_k(X,\tilde{X})$ is the} $k-$th persistence barcode of the weighted graph $\hat{\mathcal{G}}^{w, \tilde{w}}$ with the weights $w_{ij}=\lVert X_i-X_j\rVert$ and $\min(w_{ij},\tilde{w}_{ij})$, where $\tilde{w}_{ij}=\lVert \tilde{X}_i-\tilde{X}_j\rVert$.  If $\max_i \lVert X'_i-X_i\rVert=\varepsilon$, then $\lvert w'_{ij}-w_{ij}\rvert\leq 2\varepsilon$ for $w'_{ij}=\lVert X'_i-X'_j\rVert$. Similarly, $\lvert \tilde{w}'_{ij}-\tilde{w}_{ij}\rvert\leq 2\tilde{\varepsilon}$, where $\tilde{\varepsilon}=\max_j\lVert \tilde{X}'_j-\tilde{X}_j\rVert$. It follows that $\lvert\min(w'_{ij},\tilde{w}'_{ij})-\min(w_{ij},\tilde{w}_{ij})\rvert\leq 2\max(\varepsilon,\tilde{\varepsilon})$. Hence the filtration of each simplex in $VR_\alpha (\hat{\mathcal{G}}^{w, \tilde{w}})$ changes at most by $ 2\max(\varepsilon,\tilde{\varepsilon})$ under the perturbations. Next, it follows from e.g. the description of metamorphoses of canonical forms in \citep{Barannikov1994} that the birth or the death of each segment in the $k-$th barcode of $\hat{\mathcal{G}}^{w, \tilde{w}}$ changes under such perturbations at most by $2\max(\varepsilon,\tilde{\varepsilon})$. 
\end{proof}

The above arguments give also the proof for the following stability result. 
\begin{proposition}
For any quadruple of edge weights sets $w_{ij}$, ${\tilde{w}}_{ij}$, $w'_{ij}$, 
 $\tilde{w}'_{ij}$ on $\mathcal{G}$:  
%\begin{multline}\label{dbRCBw}d_B(\text{R-Cross-Barcode} _ k(w,\tilde{w}), \text{R-Cross-Barcode} _ k(v,\tilde{v})) \\ \leq \max(\max _ {ij} \lvert v _ {ij}-w _ {ij}\rvert ,\max _ {ij}\lvert \tilde{v} _ {ij}-\tilde{w} _ {ij}\rvert).\end{multline} 
%For any perturbations $w'_{ij}$  of edge weights of the graph $\mathcal{G}^{w}$ and $\tilde{w}'_{ij}$ of edge weights of the graph $\mathcal{G}^{\tilde{w}}$:
\begin{multline}\label{dbRCBw}
   d_B(\text{R-Cross-Barcode}_k(w,\tilde{w}), \text{R-Cross-Barcode}_k(w',\tilde{w}'))\\ \leq \max(\max_{ij} \lvert w'_{ij}-w_{ij}\rvert ,\max_{ij}\lvert \tilde{w}'_{ij}-\tilde{w}_{ij}\rvert) 
\end{multline} 
where $d_B$ denotes the bottleneck distance and $\text{R-Cross-Barcode}_k(w,\tilde{w})$ denotes the persistence barcode for the weighted graph $\hat{\mathcal{G}}^{w, \tilde{w}}$.
\end{proposition}
%\begin{proof}
%    Notice that $\lvert\min(w'_{ij},\tilde{w}'_{ij})-\min(w_{ij},\tilde{w}_{ij})\rvert\leq \max(\varepsilon,\tilde{\varepsilon})$. Hence the filtration of each simplex in $VR_\alpha (\hat{\mathcal{G}}^{w, \tilde{w}})$ changes at most by $ \max(\varepsilon,\tilde{\varepsilon})$ under these perturbations. It follows from e.g. the description of metamorphoses of canonical forms in \citep{Barannikov1994} that the birth or the death of each segment in the $k-$th barcode of $\hat{\mathcal{G}}^{w, \tilde{w}}$ changes under such perturbations at most by $\max(\varepsilon,\tilde{\varepsilon})$. 
%\end{proof}
\begin{proposition}
For any pair of edge weights sets ${w}_{ij}$, $\tilde{w}_{ij}$:
\begin{equation}\label{RCBww}
   \lVert \text{R-Cross-Barcode}_k(w,\tilde{w})\rVert_B \leq \max_{ij} \lvert w_{ij}-\tilde{w}_{ij}\rvert
\end{equation} where $ \lVert \rVert_B $ denotes the bottleneck norm.  
%and $\text{R-Cross-Barcode}_k(w,\tilde{w})$ denotes the persistence barcode for the weighted graph $\hat{\mathcal{G}}^{w, \tilde{w}}$ comparing the graphs $\mathcal{G}^{w}$ and $\mathcal{G}^{\tilde{w}}$.
\end{proposition}
\begin{proof} 
    Substitute $w'=\tilde{w}'=\tilde{w}$ into (\ref{dbRCBw}). 
\end{proof}
Notice that (\ref{RCBww}) is analogous to \citep[Proposition~1]{barannikov2021manifold}.

 Given  a pair of metrics $u,{u}'$ on a measure space $(\mathcal{X},\mu)$, an analogue of Gromov-Wasserstein distance between $u$ and ${u}'$ is 
\begin{equation}
  GW(u,{u}')=\inf_{e,{e}':\mathcal{X}\hookrightarrow Z}  \int_{\mathcal{X}}\rho_Z(e(x),{e}'(x))\,d\mu
\end{equation} where $e:\mathcal{X}\hookrightarrow Z$, ${e}':\mathcal{X}\hookrightarrow Z$ are  embeddings to various metric spaces $(Z,\rho_Z)$ that are isometric with respect to $u,{u}'$.
\begin{proposition} Given a triple of metrics $u,u',\tilde{u}$ on a measure space $(\mathcal{X},\mu)$,
    the expectation for the bottleneck distance between the $\text{R-Cross-Barcode}_k(w,\tilde{w})$ and the $\text{R-Cross-Barcode}_k(w',\tilde{w})$, comparing the pairs of weighted graphs associated with a sample      $X=\{x_1,\ldots, x_n\}$, $x_i\in\mathcal{X}$, 
    %$\tilde{X}=\{\tilde{x}_1,\ldots, \tilde{x}_n\}$ 
    with the edge weights $w_{ij}=u(x_i,x_j)$, $w'_{ij}=u'(x_i,x_j)$, $\tilde{w}_{ij}=\tilde{u}(x_i,x_j)$,   
    %from  $(\mathcal{X},\mu)$ 
    is upper bounded by the Gromov-Wasserstein distance between $u$ and ${u}'$: 
\begin{equation}\label{eq:dbGW}
    \int_{\mathcal{X}\times\ldots\times\mathcal{X}}d_B(\text{R-Cross-Barcode}_k(w,\tilde{w}),\text{R-Cross-Barcode}_k(w',\tilde{w})) \, d\mu^{\otimes n }\leq n \, GW(u,{u}')
\end{equation}
\end{proposition}
\begin{proof} It follows from the R-Cross-Barcode stability  (\ref{dbRCBw}) that 
\begin{multline*}
\int_{\mathcal{X}\times\ldots\times\mathcal{X}}d_B(\text{R-Cross-Barcode}_k(w,\tilde{w}),\text{R-Cross-Barcode}_k(w',\tilde{w})) \, d\mu^{\otimes n }\leq \\ \leq   \int_{\mathcal{X}\times\ldots\times\mathcal{X}} \max_{ij} \lvert w_{ij}-{w}'_{ij}\rvert    \,d\mu^{\otimes n }.
    \end{multline*}
For any pair of isometric embeddings $e:\mathcal{X}\hookrightarrow Z$, ${e}':\mathcal{X}\hookrightarrow Z$:
\begin{multline*}
    \lvert w_{ij}-{w}'_{ij}\rvert=\lvert\rho_Z(e(x _ i),{e}(x _ j)) -\rho_Z(e'(x_i),{e}'(x_j))\rvert\leq \\ \leq \rho_Z(e(x_i),{e}'(x_i)) +\rho_Z(e(x_j),{e}'(x_j))\leq \sum_{i=1}^n \rho_Z(e(x_i),{e}'(x_i))
\end{multline*} by the triangle inequality for $\rho_Z$. Therefore 
\begin{multline*}
 \int_{\mathcal{X}\times\ldots\times\mathcal{X}}d_B(\text{R-Cross-Barcode}_k(w,\tilde{w}),\text{R-Cross-Barcode}_k(w',\tilde{w})) \, d\mu^{\otimes n }\leq \\ \leq  \int_{\mathcal{X}\times\ldots\times\mathcal{X}}\sum_{i=1}^n \rho_Z(e(x_i),{e}'(x_i)) \,d\mu^{\otimes n }
    = n\, \int_{\mathcal{X}} \rho_Z(e(x),{e}'(x)) \,d\mu
\end{multline*}    
\end{proof}
\begin{proposition} The expectation for the bottleneck norm of $\text{R-Cross-Barcode} _ k(w,\tilde{w})$ for two weighted graphs with  edge weights  $w _ {ij}=u(x _ i,x _ j)$, $\tilde{w} _ {ij}=\tilde{u}(x _ i,x _ j )$, where $u ,\tilde{u}$ is a pair of metrics on a measure space $(\mathcal{X},\mu)$, and    $X  =\{x _ 1,\ldots, x _ n\}$, $x _ i\in\mathcal{X}$ is a sample from $(\mathcal{X},\mu)$,
     is upper bounded by Gromov-Wasserstein distance between $u$ and $\tilde{u}$: 
     %The expectation for the bottleneck norm of the $\text{R-Cross-Barcode}_k(w,\tilde{w})$  comparing two weighted graphs associated with a sample      $X=\{x_1,\ldots, x_n\}$, $x_i\in\mathcal{X}$, 
   % with the distance matrices  $w_{ij}=u(x_i,x_j)$, $\tilde{w}_{ij}=\tilde{u}(x_i,x_j)$,   
    %is upper bounded by the Gromov-Wasserstein distance between $u$ and $\tilde{u}$: 
\begin{equation}
    \int_{\mathcal{X}\times\ldots\times\mathcal{X}}\lVert \text{R-Cross-Barcode}_k(w,\tilde{w})\rVert_B \, d\mu^{\otimes n }\leq n \, GW(u,\tilde{u})
\end{equation}
\end{proposition}
\begin{proof}
    Substitute $u'=\tilde{u}$, $w'=\tilde{w}$ into (\ref{eq:dbGW})
\end{proof}

\section{Datasets}
\label{app:datasets}

The exact size, nature and dimension of the datasets are presented in Table \ref{tbl:dataset_description}. The errors for the synthetic data are not reported as they are zero due to the small sizes of the datasets.

\subsection{Synthetic data}

\begin{table}
    \centering
    \caption{Datasets description.}
    \label{tbl:dataset_description}
    \begin{tabular}{llll}
    \toprule
    Dataset name & Total size & Nature & Dimension  \\
    \toprule
    Circle & $1\times 10^2$ & Synthetic & $2$ \\
    \midrule
    Random & $5 \times 10^2$ & Synthetic & $2$ \\
    \midrule
    2 Clusters & $2 \times 10^2$ & Synthetic & $2$ \\
    \midrule
    3 Clusters & $3 \times 10^2$ & Synthetic & $2$ \\
    % \midrule
    % Random10D & $5 \times 10^3$ & Synthetic & $10$ \\
    \midrule
    Mammoth & $50 \times 10^3$ & Real & $3$ \\
    \midrule
    F-MNIST & $70 \times 10^3$ & Real & $784$ \\
    \midrule
    COIL-20 & $1440$ & Real & $16384$ \\
    \midrule
    scRNA mice & $1402$ & Real & $25392$ \\
    \midrule
    scRNA melanoma & $4645$ & Real & $23686$ \\
    \bottomrule
    \end{tabular}
\end{table}

The ``Random'' dataset consists of 500 points randomly distributed on a 2-dimensional unit square. The choice for this dataset was inspired by \cite{coenen2019understanding} and the ability of UMAP to find clusters in noise.

The ``Circle'' dataset is represented by 100 points randomly distributed on a 2D circle. This dataset has a simple non-trivial topology.

The ``2 Clusters'' dataset consists of 200 points, half of which goes to a dense Gaussian cluster, and the other half goes to sparse Gaussian cluster with the same mean. It is used to test the methods abilities to preserve cluster density.

The ``3 Clusters'' dataset consists of 3 Gaussian clusters each having 100 points. Two clusters are located much closer to each other than the remaining one. We propose it to test the preservation of the global structure, i.e. the distances between clusters.

\subsection{Real-world datasets}
% add short description of each dataset

Both MNIST and F-MNIST are typical datasets, consisting of 60000 28$\times$28 pixel pictures of 10 different numbers and 10 types of clothes accordingly. COIL-20 is a dataset of pictures of 20 objects taken from 72 different angles spanning 360 degrees. scRNA mice dataset has 1402 single nuclei extracted from hippocampal anatomical sub-regions (DG, CA1, CA2, and CA3), and scRNA melanoma dataset monitors expression of 4645 cells isolated from 19 metastatic melanoma patients \citep{szubert2019structure}.

Datasets licences: 
\begin{itemize}
\item Mammoth \citep{mammoth1}, CC Zero License.
Mammuthus primigenius (blumbach), The Smithsonian Institute, \textit{https://3d.si.edu/object/3d/mammuthus-primigenius-blumbach:341c96cd-f967-4540-8ed1-d3fc56d31f12}
\item MNIST \citep{lecun1998gradient}, MIT License.
\item Fashion-MNIST \citep{xiao2017online}, MIT License.
\item COIL-20 \citep{nene1996columbia}.
\item scRNA mice \citep{yuan2017challenges}.
\item scRNA melanoma \citep{tirosh2016dissecting}.
\end{itemize}

\section{More details on experiments with ``Spheres'' and ``Torus''}
\label{app:rtd_3d}

We performed experiments on dimensionality reduction to 3D space to evaluate preservation of 3-dimensional structures in data by our method. Experimental setup was  outlined in Section \ref{sec:experiments}.  

For this task we have used two synthetic datasets.

The ``Spheres'' dataset consists of 17,250 points randomly distributed on surface of eleven 100-spheres in 101-dimensional space. Any two of those do not intersect and one of the spheres contains all other inside. Similar to ``Circle'' dataset (Section \ref{subsec:2d2d}) UMAP splits bigger sphere (light grey) into 10 parts and wraps each small sphere into one of them. PacMAP performs similar but it also splits a part of bigger sphere into separate sphere. PCA and Ivis preserve the shape of inner spheres only and turn all structure `inside out'. Both t-SNE and regular AE projects all points onto one sphere without clear separation between clouds. The addition of a topological loss, both in TopoAE and in our RTD-AE, preserves the global structure of inlaid clusters. However, TopoAE flattens inner clusters into disks, while RTD-AE makes them into (hollow) spheres.

\begin{figure}[tbp]
    \centering
    \includegraphics[width=\textwidth]{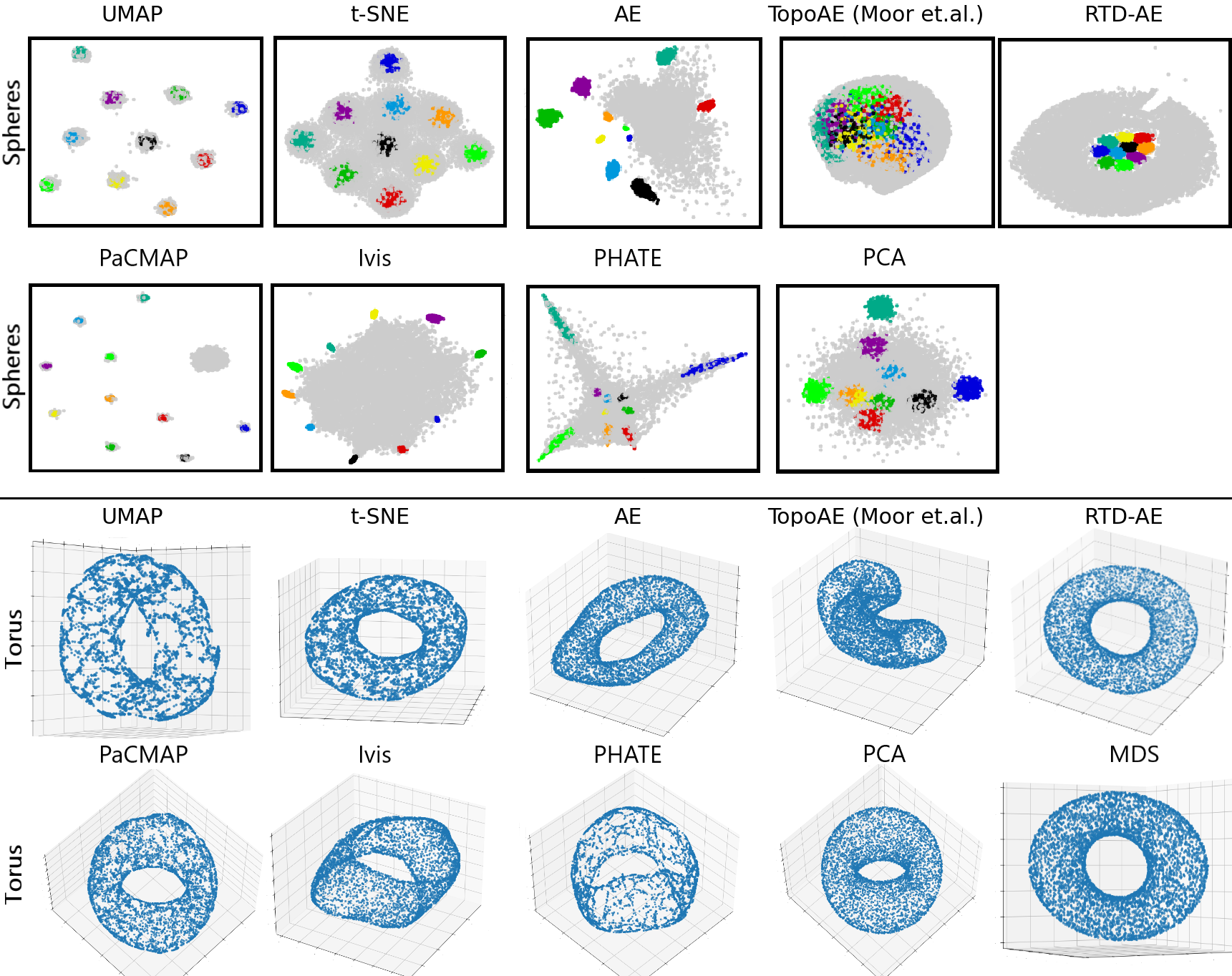}
    \caption{Results on dimensionality reduction of Spheres to 2D-space and Torus to 3D-space}
    \label{fig:synthetic3d_appn}
\end{figure}

The ``Torus'' dataset consists of 5,000 points randomly distributed on surface of a 2-torus ($T^2$) immersed into 100-dimensional space. Due to such nature of this dataset, PCA and MDS methods perform on it very well. RTD-AE takes the third place with very similar quality, see Figure \ref{fig:synthetic3d_appn} and Table \ref{tab:3D2D}.

For ``Spheres'' dataset we have also performed experiments on dimensionality reduction to 2D space. Overall results are quite similar to those obtained for 3D case. The behavior of baselines remains essentially the same. The only interesting change is that RTD-AE now projects bigger sphere to a ring  and puts the projections  of smaller spheres into the ring's hollow center. RTD-AE outperforms other methods in terms of linear correlation and triplet accuracy. 
%Metrics presented at Table show PCA and MDS are clearly winning in terms of Linear Correlation of distances with RTD-AE slightly behind those two, but later is clearly winning when it comes to other two metrics -- Wasserstein distance and triplet accuracy.

All of the representations were generated with default parameters of baseline methods. 
Results are presented at Figures \ref{fig:synthetic3d} (``Spheres'' to 3D space) and \ref{fig:synthetic3d_appn} (``Spheres'' to 2D space and ``Torus'' to 3D). 

\begin{table}[t]
  \caption{Quality of data manifold global structure preservation at projection Torus dataset from 100D into 3D space and Spheres dataset from 101D to 2D.}
  \centering
  \begin{tabular}{lllllll}
    \toprule
    & & \multicolumn{5}{c}{Quality measure}           \\
    \cmidrule(r){3-7}
    Dataset & Method   & L. C. & W. D. $H_{0}$ & W. D. $H_{1}$ & T. A. & RTD \\
    \midrule
    Torus 2D & t-SNE & {0.989} & {1.021} $\pm$ 0.07 &  {0.594} $\pm$ 0.05 & {0.896} $\pm$ 0.01 &  {1.533} $\pm$ 0.09 \\
        & UMAP & 0.955 & 2.052 $\pm$ 0.15 & {0.931} $\pm$ 0.07  & 0.931 $\pm$ 0.07 & {3.250} $\pm$ 0.17 \\
        & PaCMAP & 0.987 & 1.410 $\pm$ 0.12 & 0.833 $\pm$ 0.08 & 0.883 $\pm$ 0.01 & 2.114 $\pm$ 0.08 \\
        & PHATE & 0.873 & 2.967 $\pm$ 0.33 & 1.143 $\pm$ 0.09 & 0.646 $\pm$ 0.02 & 4.061 $\pm$ 0.20 \\
        & PCA & \textbf{1.0} & \textbf{0.871 $\pm$ 0.24} & \textbf{0.014 $\pm$ 0.00} & \textbf{0.999 $\pm$ 0.00} &\textbf{0.000 $\pm$ 0.00} \\
        & MDS & \textbf{1.0} & \textbf{0.880 $\pm$ 0.24} & \underline{0.022} $\pm$ 0.00 & \textbf{0.999 $\pm$ 0.00} & \textbf{0.000 $\pm$ 0.00} \\
        & Ivis & 0.844 & 2.606 $\pm$ 0.27 & 1.086 $\pm$ 0.11 & 0.580 $\pm$ 0.02 & 4.073 $\pm$ 0.17 \\
        & AE & 0.880 & 2.023 $\pm$ 0.30 & 1.073 $\pm$ 0.08 & 0.662 $\pm$ 0.02 & 3.433 $\pm$ 0.12 \\
        & TopoAE & 0.920 & 2.616 $\pm$ 0.34 & 1.017 $\pm$ 0.09 & 0.696 $\pm$ 0.02 & 2.975  $\pm$ 0.14 \\
        & RTD-AE & \underline{0.992} & \underline{0.907} $\pm$ 0.08 & 0.109 $\pm$ 0.01 & \underline{0.902} $\pm$ 0.01 & \underline{0.148} $\pm$ 0.01 \\
    \midrule
     Spheres 2D & t-SNE & {0.018} & 49.77 $\pm$ 1.40 & 0.349 $\pm$ 0.05 & 0.166 $\pm$ 0.01 & 44.00 $\pm$ 1.44\\
            & UMAP & 0.020 & \underline{47.55} $\pm$ 1.33 & 0.233 $\pm$ 0.03 & 0.191 $\pm$ 0.01 & 45.41 $\pm$ 1.47 \\
            & PaCMAP & \underline{0.342} & \textbf{46.57 $\pm$ 1.68}  & \underline{0.208} $\pm$ 0.02 & 0.155 $\pm$ 0.01 & 45.56 $\pm$ 1.46 \\
            & PHATE & 0.040 & 48.68 $\pm$  1.70 & \textbf{0.188 $\pm$ 0.03} & 0.201 $\pm$ 0.01 & 45.08 $\pm$ 1.93\\
            & PCA & 0.117 & 49.58 $\pm$ 1.60 & 0.447 $\pm$ 0.05 & 0.180 $\pm$ 0.02 & {43.01} $\pm$ 1.36 \\
            & MDS & N.A. & N.A. & N.A.& N.A. & N.A.\\
            & Ivis & 0.280 & 48.84 $\pm$ 1.73 & 0.342 $\pm$ 0.05 & 0.125 $\pm$ 0.01 & 44.21 $\pm$ 1.36\\
            & AE & 0.334 & {48.31 $\pm$ 1.74}  & 0.320 $\pm$ 0.04  & 0.124 $\pm$ 0.01 & 43.74 $\pm$ 1.60 \\
            & TopoAE & {0.264} & 49.94 $\pm$ 1.52 & 0.634 $\pm$ 0.06 & \underline{0.245} $\pm$ 0.02 & \underline{42.70} $\pm$ 1.74 \\
            & RTD-AE & \textbf{0.611} & {48.20 $\pm$ 1.72} & 0.538 $\pm$ 0.05 &\textbf{0.343 $\pm$ 0.01}  & \textbf{41.22 $\pm$ 1.70}\\
        
    \bottomrule
  \end{tabular}\label{tab:3D2D}
\end{table}

% \begin{figure}[!h]
%     \centering
%     \includegraphics[width=\textwidth]{img/coil20_modifications.png}
%     \caption{Modifications of RTD-AE on COIL-20 dataset.}
%     \label{fig:coil20_mod}
% \end{figure}

%\newpage

\begin{table}[t]
  \caption{Quality of data manifold global structure preservation for projection of COIL-20 into 3D-space.}
  \label{tbl:app_measure_synthetic3d}
  \centering
  \begin{tabular}{lllll}
    \toprule
    & & \multicolumn{3}{c}{Quality measure}           \\
    \cmidrule(r){3-5}
    Dataset & Method & L. C. & W. D. $H_{0}$ & T. A. \\
    \midrule
%     Spheres & t-SNE & {0.087} & 47.89 $\pm$ 2.59 & 1.376 $\pm$ 0.11  & 0.206 $\pm$ 0.01 & 0.573 \\
%             & UMAP & 0.049 & 48.31 $\pm$ 1.83 & \textbf{0.269 $\pm$ 0.03} & 0.192 $\pm$ 0.01 & \textbf{0.602} \\
%             & AE & 0.441 & \underline{45.07 $\pm$ 2.27} & 1.130 $\pm$ 0.11 & 0.333 $\pm$ 0.02 & 0.565 \\
%             & TopoAE & {0.424} & 45.89 $\pm$ 2.35 & 1.410 $\pm$ 0.12 & 0.274 $\pm$ 0.02 & 0.590 \\
%           & RTD-AE & \textbf{0.633} & \textbf{45.02 $\pm$ 2.69} & \underline{0.956 $\pm$ 0.10} & \textbf{0.346 $\pm$ 0.02} & 0.565 \\
%             & RTD-g  & {0.?} & {0.?} & {0.?} & 0.? & \\
%             & RTD-gMM & 0.? & 0.? & 0.? & 0.? & \\
%             & RTD-gL2  & 0.? & 0.? & 0.? & 0.? & \\
%     \midrule
%   Torus & t-SNE & \underline{0.989} & \underline{1.021} $\pm$ 0.07 & \underline{0.608} $\pm$ 0.05 & \underline{0.896} $\pm$ 0.01 & N/A \\
%       & UMAP & 0.955 & 2.052 $\pm$ 0.15  & 0.904 $\pm$ 0.09 & 0.847 $\pm$ 0.01 & N/A \\
%         & AE & 0.880 & 2.023 $\pm$ 0.30 & 1.012 $\pm$ 0.09 & 0.662 $\pm$ 0.02 & N/A \\
%         & TopoAE & 0.920 & 2.616 $\pm$ 0.34 & 0.953 $\pm$ 0.14 & 0.696 $\pm$ 0.02 & N/A \\
%         & RTD-AE & \textbf{0.992} & \textbf{0.907} $\pm$ 0.08 & \textbf{0.175} $\pm$ 0.02 & \textbf{0.902} $\pm$ 0.01 & N/A\\
%         & RTD-g  & \textbf{0.?} & 0.? & 0.? & N/A & N/A \\
%         & RTD-gMM & \textbf{0.?} & \textbf{0.?} & \textbf{0.?} & N/A & N/A \\
%         & RTD-gL2  & 0.? & 0.? & 0.? & N/A & N/A \\
%     \midrule
    COIL-20 & t-SNE & 0.608 & 255 $\pm$ 0.0 & 0.706 $\pm$ 0.01 \\
        & UMAP & 0.250 & 278 $\pm$ 0.0  & 0.574 $\pm$ 0.012 \\
        & AE & 0.792 & 253 $\pm$ 0.0  & 0.803 $\pm$ 0.009 \\
        & TopoAE & 0.677 & 236 $\pm$ 0.0  & 0.740 $\pm$ 0.016 \\
        & RTD-AE & \textbf{0.811} & \textbf{233 $\pm$ 0.0}  & \textbf{0.814 $\pm$ 0.014} \\
        % & RTD-AE-MM & \textbf{0.816} & 238 $\pm$ 0.0 & \textbf{0.829} $\pm$ 0.009 \\
    \bottomrule
  \end{tabular}
\end{table}

\section{Ablation study}
\begin{figure}
    \centering
    \includegraphics[width=\linewidth]{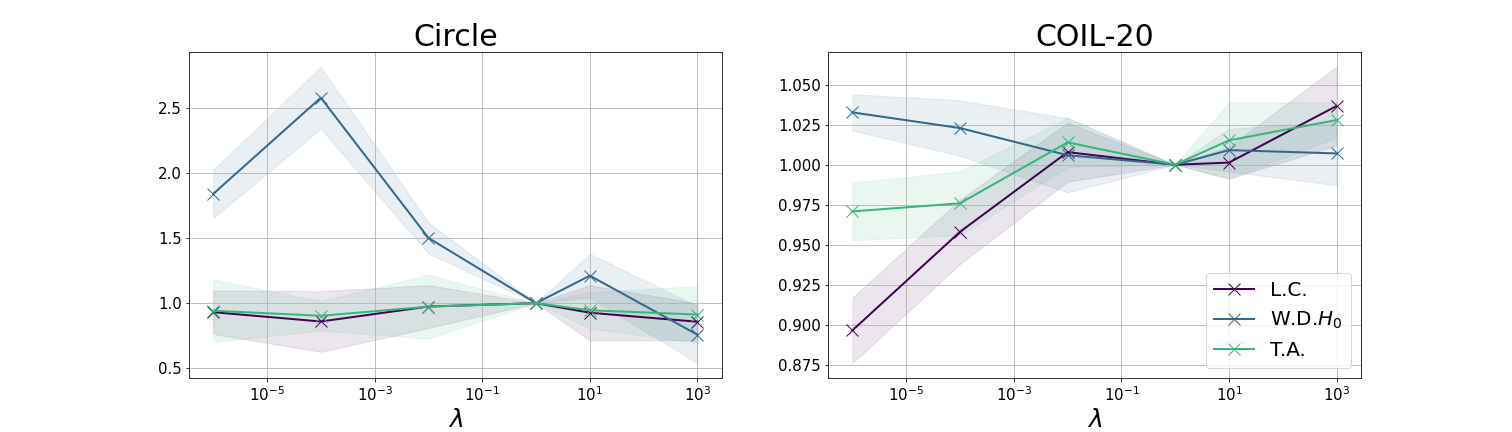}
    \caption{Results of ablation study. The plot depicts the value of the metrics relative to its value at $\lambda=1.0$, which was used in all previous experiments. We clearly see that the addition of our RTD loss indeed increases linear correlation and triplet accuracy and at the same time decreases W.D.$H_0$. At the same time choosing $\lambda = 1.0$ seems reasonable to us as increasing its value further does not affect on the quality.}
    \label{fig:ablation}
\end{figure}

In this section we investigate the effect of adding RTD loss on the performance of the model. We add a hyperparameter $\lambda$ responsible for the scale of the RTD loss variable: $\mathcal{L}_{rec}(X, \tilde{X}) + \lambda \text{RTD}(X, Z)$. We run our experiments on two datasets: COIL-20 and Circle. The hyperparameter value ranged from $10^{-6}$ to $10^{3}$. For each value of $\lambda$ we run the procedure 8 times to get the confidence levels of our metrics. The results are depicted at Figure \ref{fig:ablation}.

\section{More dimensionality reduction methods on ``Mammoth'' dataset} 

\begin{figure}[t]
    \centering
    \begin{subfigure}{0.35\textwidth}
    \includegraphics[width=\textwidth]{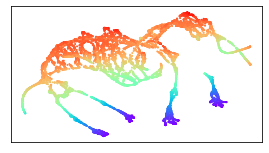}
    \caption{PaCMAP}
    \end{subfigure}
    \begin{subfigure}{0.35\textwidth}
    \includegraphics[width=\textwidth]{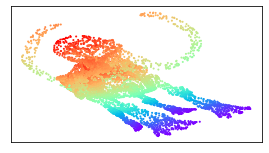}
    \caption{Ivis}
    \end{subfigure}
    \caption{Additional dimensionality reduction methods applied to the ``Mammoth'' dataset}
    \label{fig:mammoth_pacmap_ivis}
\end{figure}

See Figure \ref{fig:mammoth_pacmap_ivis}.

\section{R-Cross-Barcodes}
\label{app:r-cross-barcodes}

\begin{figure}[t]
    \centering
    \includegraphics[width=\textwidth]{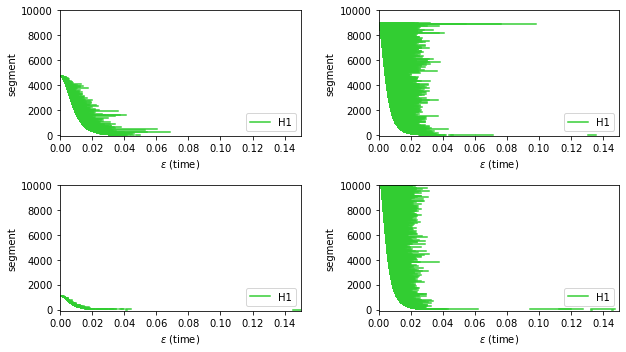}
    \caption{R-Cross-Barcodes between latent representations and original data points. Top: R-Cross-Barcode$(Z_{0}, X)$, R-Cross-Barcode$(X, Z_{0})$. Bottom: R-Cross-Barcode$(Z, X)$, R-Cross-Barcode$(X, Z)$.
    $X$ - ``Mammoth`` dataset, $Z$ - latent representations from RTD-AE, $Z_{0}$ - latent representation from the untrained autoencoder. Intervals in R-Cross-Barcodes are smaller after training.}
    \label{fig:r_cross_barcodes}
\end{figure}

See Figure \ref{fig:r_cross_barcodes}.

\begin{figure}[tbh!]
    \centering
    \begin{subfigure}{0.35\textwidth}
    \includegraphics[width=\textwidth]{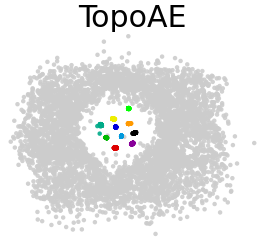}
  % \caption{Original 3D data}
    \end{subfigure}
    \begin{subfigure}{0.35\textwidth}
     \includegraphics[width=\textwidth]{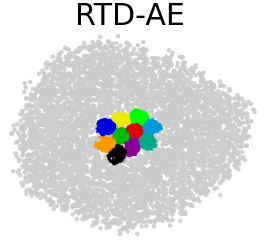}
  %\caption{RTD}
    \end{subfigure}
    \caption{{Dimensionality reduction of Spheres dataset to 2D-space after hyperparameter search.}}
  % Notice a substantial change for the TopoAE representation, while the RTD-AE representation essentially does not change.
    \label{fig:spheres2Dtuned}
\end{figure}

\begin{figure}[t]
    \centering
    \begin{subfigure}{0.35\textwidth}
    \includegraphics[width=\textwidth]{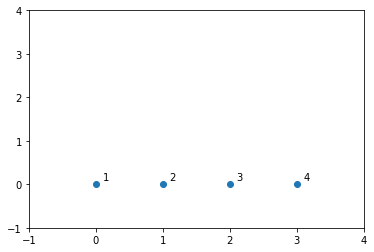}
    \caption{The point cloud $X$.}
    \end{subfigure}
    \begin{subfigure}{0.35\textwidth}
    \includegraphics[width=\textwidth]{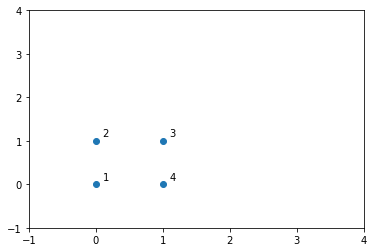}
    \caption{The point cloud $\tilde{X}$.}
    \end{subfigure}
    \\
    \begin{subfigure}{0.35\textwidth}
    \includegraphics[width=\textwidth]{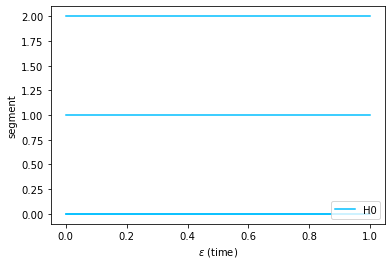}
    \caption{The barcode of $X$.}
   \end{subfigure}
    \begin{subfigure}{0.35\textwidth}
    \includegraphics[width=\textwidth]{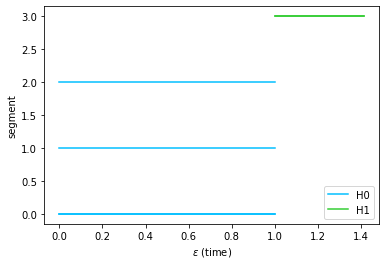}
    \caption{The barcode of $\tilde{X}$.}
    \end{subfigure}
    \\
    \begin{subfigure}{0.35\textwidth}
    \includegraphics[width=\textwidth]{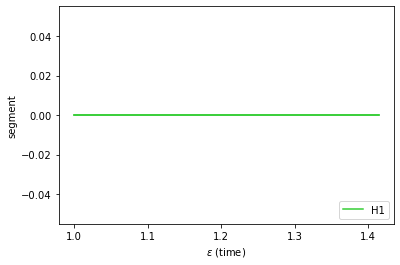}
    \caption{The R-Cross-Barcode$_1(X,\tilde{X})$}
    \end{subfigure}
    %\begin{subfigure}{0.35\textwidth}
    %\includegraphics[width=\textwidth]{img/topoae_bug_rcbarcode2.png}
     %\caption{The R-Cross-Barcode$_1(\tilde{X}, X)$}
    %\end{subfigure}

    \caption{{Two point clouds \mbox{$X, \tilde{X}$} for which the identity of indiscernibles property doesn't hold for the topological term in the TopoAE loss. The one-to-one correspondence between clouds is depicted by numbers. The minimal spanning trees $1-2-3-4$ have edges of identical length for both point clouds.  For these point clouds,
\mbox{$\text{RTD}(X, \tilde{X})=0.207$}, while the topological term of the TopoAE loss equals $0$. The topology of these point clouds is different, in particular they have different barcodes.   The distinguishing topological feature between $X$ and $\tilde{X}$ is the cycle in $\tilde{X}$ which is born at $\alpha=1$ and dies at $\alpha=\sqrt{2}$. The R-Cross-Barcode$_1(\tilde{X}, X)$ depicts this difference. } }
    \label{fig:topoae_bug}
\end{figure}

\section{Reconstruction loss}

{See Table} \ref{tbl:app_reconstruction_loss} {for results.}

\begin{table}[t]
  \caption{{Reconstruction loss for when projecting into 16 dimension latent space}}.
  \label{tbl:app_reconstruction_loss}
  \centering
  \begin{tabular}{lll}
    \toprule
    % & & \multicolumn{3}{c}{Quality measure}           \\
    % \cmidrule(r){3-5}
    Dataset & Method & Reconstruction loss \\
    \midrule
    COIL-20 & AE & $1.89 \times 10^{-4}$\\
    & TopoAE & $3.30 \times 10^{-4}$\\
    & RTD-AE & $4.54 \times 10^{-4}$\\
    \midrule
    F-MNIST & AE & $2.73 \times 10^{-3}$\\
    & TopoAE & $2.84 \times 10^{-3}$\\
    & RTD-AE & $3.17 \times 10^{-3}$\\
    \midrule
    MNIST & AE & $3.78 \times 10^{-3}$\\
    & TopoAE & $3.70 \times 10^{-3}$\\
    & RTD-AE & $4.88 \times 10^{-3}$\\
    \midrule
    scRNA mice & AE & $1.31 \times 10^{-3}$\\
    & TopoAE & $1.23 \times 10^{-3}$\\
    & RTD-AE & $1.32 \times 10^{-3}$\\
    \midrule
    scRNA melanoma & AE & $1.16 \times 10^{-3}$\\
    & TopoAE & $1.11 \times 10^{-3}$\\
    & RTD-AE & $1.15 \times 10^{-3}$\\
    \bottomrule
    \end{tabular}
\end{table}

\section{Hyperparameters search for Spheres dataset (into 2D) }

{For TopoAE we performed hyperparametrs search in accordance with the original paper \mbox{\cite{moor2020topological} and selected best combination according to KL$_{0.1}$-divergence.}}
%Additionally we provide KL$_{0.01}$ and KL$_{1.0}$.

{For RTD-AE we searched for batch size in \mbox{ $[20; 250]$ and $\lambda$ in $[0.1; 10]$}. Best combination was once again selected w.r.t. \mbox{KL$_{0.1}$-divergence.} }

{Results are presented in \mbox{Table \ref{tbl:2dspheres_finetuning}.}
For Wasserstein Distance and Triplet Accuracy difference between means is lesser than standard derivations, and due to this, we performed one-tailed Student's t-test to verify their relation. According to its results, we can reject the null hypothesis that the mean W.D. $H_0$ for TopoAE is lower than the mean W.D. $H_0$ for RTD-AE at a significance level of 0.05. Same result confirming the better performance of RTD-AE was obtained for the triplet accuracy.}
 
\begin{table}[t]
  \caption{{Hyperparameter search for Spheres (into 2D) dataset}}.
  \label{tbl:2dspheres_finetuning}
  \centering
  \begin{tabular}{lllllll}
    \toprule
    % & & \multicolumn{3}{c}{Quality measure}           \\
    % \cmidrule(r){3-5}
    Dataset & Method &L.C. & W.D. $H_0$  & T.A. & RTD \\
%& KL$_{0.001}$& KL$_{0.01}$ & KL$_{1.0}$\\
    \midrule
    Spheres 2D  & TopoAE & $0.691$ & $43.291 \pm 1.583 $ &  $0.3688 \pm 0.0165 $ & $39.837 \pm 1.318 $ \\
%& $0.1270$ & $0.2976$ & $0.0057$\\
    & RTD-AE & $\mathbf{0.706}$ & $\mathbf{42.133 \pm 1.683} $ & $\mathbf{0.3765 \pm 0.0124}$ & $\mathbf{37.286 \pm 1.393}$ \\
%& $0.0612$ & $0.2482$ & $0.0047$ \\
    \bottomrule
    \end{tabular}
\end{table}

\section{Identity of indiscernibles for the TopoAE loss}
\label{append:indescTopoAE}
{
We compare two point clouds $X, \tilde{X}$ from Figure \mbox{\ref{fig:topoae_bug}. For these point clouds,
$\text{RTD}(X, \tilde{X})=0.207$}, while the topological part of the TopoAE loss equals $0$. The distinguishing topological feature between $X$ and $\tilde{X}$ is the cycle in $\tilde{X}$ which is born at $\alpha=1$ and dies at $\alpha=\sqrt{2}$. R-Cross-Barcode($\tilde{X}$, X) depicts this difference.
}

\end{document}